\documentclass[journal]{IEEEtran}
\ifCLASSINFOpdf
\else
\fi

\usepackage{graphicx}
\usepackage{caption}

\usepackage{cite}
\usepackage{comment}
\usepackage{amsmath,amssymb} 
\usepackage{xcolor}
\usepackage[normalem]{ulem}
\usepackage{multirow}
\usepackage{threeparttable}
\usepackage{booktabs}
\usepackage{floatrow}
\newfloatcommand{capbtabbox}{table}[][\FBwidth]
\usepackage{mathtools}
\usepackage{enumitem}

\def\figvspace{{\vspace{-3mm}}}
\def\intervspace{{\vspace{1mm}}}

\begin{document}
%
\title{Learning for Unconstrained Space-Time Video Super-Resolution}
%
%
%

\author{Zhihao Shi,~
		Xiaohong Liu,~\IEEEmembership{Graduate Student~Member,~IEEE},~
		Chengqi Li,~
		Linhui Dai,~
		Jun Chen,~\IEEEmembership{Senior Member,~IEEE},~
		Timothy N. Davidson,~\IEEEmembership{Fellow,~IEEE},~
		Jiying Zhao,~\IEEEmembership{Member,~IEEE}
}

\maketitle

\newcommand\blfootnote[1]{%
	\begingroup
	\renewcommand\thefootnote{}\footnote{#1}%
	\addtocounter{footnote}{-1}%
	\endgroup
}

\begin{figure*}[t]
	\centering
	\begin{minipage}[htb]{0.16\linewidth}
		\centering
		{\includegraphics[width=\linewidth]{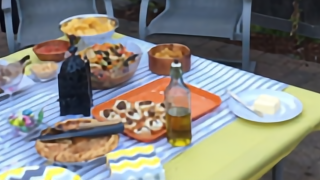}}
		{\includegraphics[width=\linewidth]{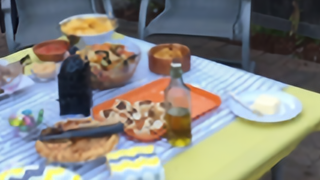}}
		\centerline{$s=3, t=0.250$}
		\vspace{0.0001cm}
	\end{minipage}
	\begin{minipage}[htb]{0.16\linewidth}
		\centering
		{\includegraphics[width=\linewidth]{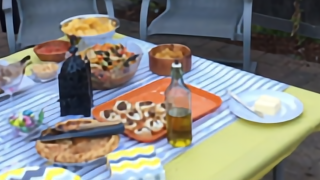}}
		{\includegraphics[width=\linewidth]{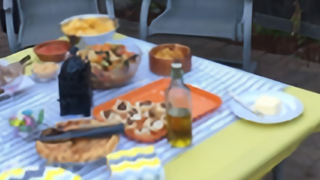}}
		\centerline{$s=3, t=0.500$}
		\vspace{0.0001cm}
	\end{minipage}
	\begin{minipage}[htb]{0.16\linewidth}
		\centering
		{\includegraphics[width=\linewidth]{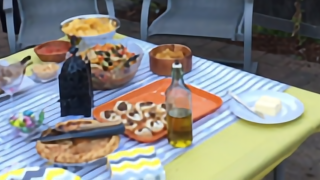}}
		{\includegraphics[width=\linewidth]{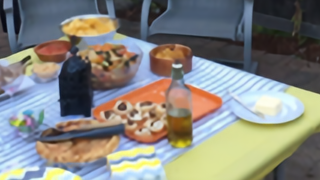}}
		\centerline{$s=3, t=0.625$}
		\vspace{0.0001cm}
	\end{minipage}
	\begin{minipage}[htb]{0.16\linewidth}
		\centering
		{\includegraphics[width=\linewidth]{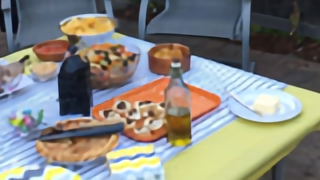}}
		{\includegraphics[width=\linewidth]{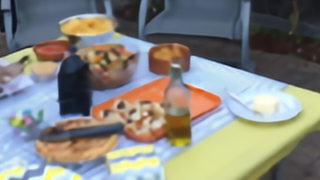}}
		\centerline{$s=4, t=0.250$}
		\vspace{0.0001cm}
	\end{minipage}
	\begin{minipage}[htb]{0.16\linewidth}
		\centering
		{\includegraphics[width=\linewidth]{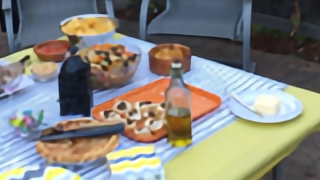}}
		{\includegraphics[width=\linewidth]{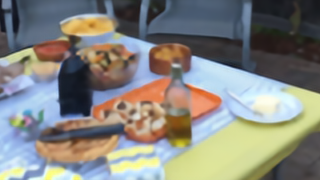}}
		\centerline{$s=4, t=0.500$}
		\vspace{0.0001cm}
	\end{minipage}
	\begin{minipage}[htb]{0.16\linewidth}
		\centering
		{\includegraphics[width=\linewidth]{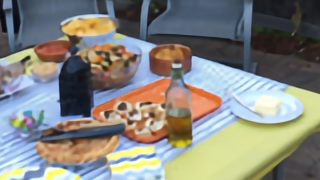}}
		{\includegraphics[width=\linewidth]{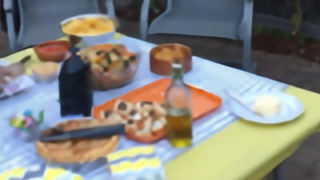}}
		\centerline{$s=4, t=0.625$}
		\vspace{0.0001cm}
	\end{minipage}
	\figvspace
	\caption{Comparison between the proposed method (the first row) and a state-of-the-art two-stage method: BMBC \cite{park2020bmbc} + Meta-SR \cite{hu2019metasr} (the second row).}
	\label{fig:start show}
\end{figure*}

\begin{abstract}
Recent years have seen considerable research activities devoted to video enhancement that simultaneously increases temporal frame rate and spatial resolution. However, the existing methods either fail to explore the intrinsic relationship between temporal and spatial information or lack flexibility in the choice of final temporal/spatial resolution. In this work, we propose an unconstrained space-time video super-resolution network, which can effectively exploit space-time correlation to boost performance. Moreover, it has complete freedom in adjusting the temporal frame rate and spatial resolution through the use of the optical flow technique and a generalized pixelshuffle operation. Our extensive experiments demonstrate that the proposed method not only outperforms the state-of-the-art, but also requires far fewer parameters and less running time.
\end{abstract}


\blfootnote{Z.~Shi, X.~Liu (corresponding author), C.~Li, L.~Dai, J.~Chen and T.N. Davidson are with the Department of Electrical and Computer Engineering, McMaster University, Hamilton, ON., L8S 4K1, Canada (e-mail: \{shiz31, liux173, lic222, dail5, chenjun, davidson\}@mcmaster.ca).
	
J.~Zhao is with the School of Electrical Engineering and Computer Science, University of Ottawa, Ottawa, ON., K1N 6N5, Canada (e-mail: jzhao@uottawa.ca)

This work was supported in part by the Natural Sciences and Engineering Research Council of Canada
through a Discovery Grant.}

\vspace{-1mm}

%
\IEEEpeerreviewmaketitle

\section{Introduction}
Recently, we have witnessed the popularization of Ultra-High-Definition TeleVision (UHDTV) and the rising of UHD TV shows in broadcasting. However, despite new media contents can be filmed by the advanced UHD recorder, remaking a large quantity of existed ones is impractical, leading to the overall short supply. Video Super-Resolution (VSR) technologies provide a promising way to reconstruct High-Resolution (HR) videos from their Low-Resolution (LR) counterparts. Furthermore, while watching sport events on TV, one may playback the fleeting moments with slow motion. Video Frame Interpolation (VFI) is one of the solutions that can temporally increase the frame rate of the broadcast videos. 

In this paper, Space-Time Video Super-Resolution (STVSR), the combination of VSR \cite{zhang2021video, wang2019edvr, haris2019recurrent, yi2019progressive, tian2020tdan, isobe2020video, liu2021exploit} and VFI \cite{wang2010motion, yang2008new,yan2020fine, xue2019video, bao2019memc, bao2019depth, jiang2018super, niklaus2018context, qvi_nips19, lee2020adacof, niklaus2017video, niklaus2018video}, is mainly researched that aims at increasing spatial resolution and temporal frame rate simultaneously. The traditional approaches to STVSR \cite{shechtman2005space, mudenagudi2010space, takeda2010spatiotemporal, shahar2011space} typically rely on strong assumptions or hand-crafted priors, and consequently are only suited to specific scenarios. 
The advent of deep learning has revolutionized many areas in computer vision, including, among others, image super-resolution \cite{jiang2020lightweight, wan2020lightweight}, image quality assessment \cite{qu2021light}, image deblurring \cite{esmaeilzehi2021updresnn}, image compression \cite{tian2020just}, and video coding \cite{lei2021deep}.
In particular, it enables the development of data-driven approaches to VFI and Super-Resolution (SR) that can capitalize on the learning capability of neural networks as opposed to relying on prescribed rules. STVSR also naturally benefits from this advancement since it can be realized via a direct combination of VFI and SR. Specifically, one can first use VFI to increase the temporal frame rate, then leverage SR to enhance the spatial resolution. Moreover, the State-Of-The-Art (SOTA) VFI and SR methods (\textit{e.g.}, the flow-based VFI methods \cite{xue2019video, bao2019memc, bao2019depth, jiang2018super, niklaus2018context, qvi_nips19} and the meta-learning-based SR methods \cite{hu2019metasr}) have the freedom to adjust the frame rate and the spatial resolution, respectively. As a consequence, the resulting two-stage scheme is able to perform unconstrained STVSR. 
However, as pointed out in \cite{haris2020space, xiang2020zooming}, this two-stage scheme does not take advantage of the intrinsic relationship between temporal and spatial information, which limits the highest resolution that can be potentially achieved (see Fig.~\ref{fig:start show}). In addition, performing STVSR in a two-stage fashion tends to be highly inefficient since VFI and SR are computationally intensive by themselves and likely involve many operations that can be shared.

To tackle these problems, two recent works \cite{haris2020space, xiang2020zooming} have proposed a one-stage approach to STVSR by consolidating VFI and SR. This boosts performance by a large margin, while involving far fewer parameters and incurring less computational cost. However, this gain comes at a price. Indeed, compared to its two-stage counterpart, the new approach in \cite{haris2020space, xiang2020zooming} lacks flexibility in the choice of the final temporal/spatial resolution. Specifically, in the temporal domain, the Convolution Neural Network (CNN) employed to synthesize the intermediate frame (based on two input frames) is tailored to a particular target time. As for the spatial domain, due to the use of the pixelshuffle layer \cite{shi2016real} or deconvolution layer, it is impossible to adjust the up-sampling factor without modifying or retraining the network. Besides, the intrinsic limitation of these two layers renders fractional up-sampling factors unrealizable.

A natural question that arises here is whether the performance of the one-stage scheme can be retained without compromising flexibility?
We offer an affirmative answer in this work by proposing an Unconstrained Space-Time Video Super-Resolution Network (USTVSRNet), which is able to increase the temporal/spatial resolution of a given video by an arbitrary factor. For temporal interpolation, the optical flow technique is adopted to ensure the desired flexibility in the temporal resolution. Moreover, different from \cite{haris2020space, xiang2020zooming}, where the intermediate frame is synthesized at the feature level, we make predictions at both the image and feature levels, which leads to a noticeable performance improvement. As to spatial up-sampling, we introduce a Generalized Pixelshuffle Layer (GPL) that can project low-dimensional features to a high-dimensional space with the dimension ratio freely chosen. In addition, we construct a Scale-Attentive Residual Dense Block (SARDB) to generate scale-aware features. Due to the innovative features of our design, USTVSRNet is capable of up-sampling frames by an arbitrary factor with a single model. Our experimental results will show that the proposed method outperforms the SOTA two-stage methods, and does so with significantly lower computational cost.

The main contributions of this paper are as follows: (1) We propose a novel unconstrained STVSR method, which possesses the strengths of the SOTA one-stage and two-stage approaches while avoiding their drawbacks. (2) To realize unconstrained STVSR, several new mechanisms are introduced, including, integrating image-level and feature-level information to improve the quality of the synthesized intermediate frame, generalizing the Standard Pixelshuffle Layer (SPL) to increase the degrees of freedom in terms of up-sampling factor, and generating scale-aware features to make the network more adaptive. (3) Even when evaluated for particular temporal/spatial resolutions, the performance of the proposed unconstrained STVSR method remains highly competitive and outperforms the SOTA one-stage methods on various datasets.

\section{Related Work}


\subsection{Video Frame Interpolation} The goal of VFI is to increase the frame rate by synthesizing intermediate frames while maintaining spatial and temporal consistencies with the given video frames. There are two major categories of video interpolation methods: kernel-based and flow-based methods.

As a pioneer of the kernel-based method, reference \cite{niklaus2017video} employs a rigid spatially-adaptive convolution kernel to generate each target pixel. Naturally, very large kernels are needed for covering large motions, which leads to a substantial memory overhead. Reference \cite{niklaus2018video} replaces regular 2D convolution kernels with pairs of 1D kernels to reduce the memory overhead. Even though that reduction is significant, the method cannot handle motions that are larger than the kernel size. To solve this problem, AdaCoF \cite{lee2020adacof} breaks the rigid limitation of the regular convolution kernel and proposes a 2D deformable spatially-adaptive convolution scheme for VFI. Later, GDConvNet, introduced in \cite{shi2020video}, further exploited the degrees of freedom available in the three dimension of space-time, which improves the performance significantly.
While kernel-based methods show promise, the time-oblivious nature of the convolution kernels means that the temporal information in the intermediate frames needs to be built into kernel-based methods in the design phase and cannot be easily adjusted during implementation.

In contrast, flow-based methods \cite{jiang2018super, bao2019depth, bao2019memc, qvi_nips19, niklaus2018context} generate the value of each pixel in the
target intermediate frame according to an associated optical flow. Specifically, they first use the input frames to estimate source optical flows with the help of an optical flow estimation network \cite{dosovitskiy2015flownet, ranjan2017optical,ilg2017flownet,sun2018pwc}. 
They then convert the source optical flows into the associated ones with respect to the intermediate time $t$. Finally, the input frames are warped to the target frame according to these optical flows. As such, these methods have the inherent ability to perform interpolation with respect to an arbitrary time. Flow-based methods \cite{jiang2018super, bao2019depth, bao2019memc, niklaus2018context} typically adopt a linear model to convert the source optical flows.
Recently, a quadratic model was proposed in \cite{qvi_nips19}, and preliminary results suggest that it may be able to better estimate the optical flows by exploiting four consecutive frames.
For simplicity, in the present paper we will focus on the linear model, which involves two consecutive frames. It is straightforward to extend our work to incorporate higher-order models.


\begin{figure*}[t]
	\centering
	\begin{minipage}[h]{\linewidth}
		\centering
		\includegraphics[width=\linewidth]{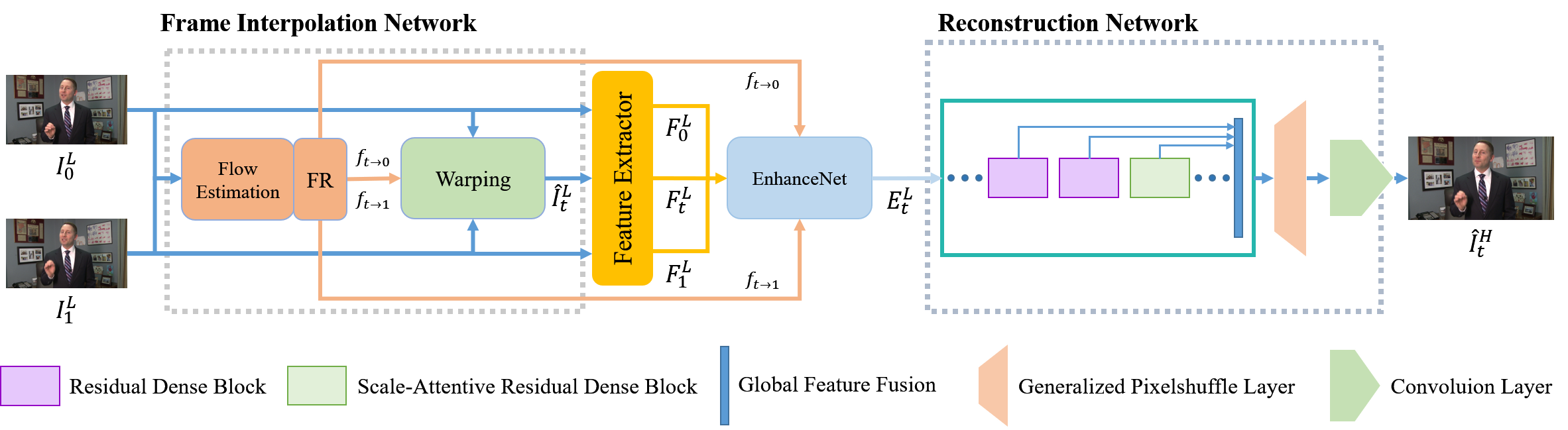}
	\end{minipage}
	\figvspace
	\caption{Illustration of the architecture of USTVSRNet.}

	\label{fig:FrameWork}
\end{figure*}

\subsection{Super Resolution}
SR has two main branches: Single Image Super-Resolution (SISR) and video super-resolution, which aim at recovering a visually pleasing high-resolution image and video, respectively.

In terms of SISR,
an end-to-end network which maps the interpolated LR images to HR ones was proposed in \cite{dong2014learning}, and was enhanced by increasing network depth or stacking more complicated modules in \cite{kim2016accurate, zhang2017learning, haris2018deep, zhang2018residual}. However, all of these methods need to pre-compute an interpolated LR image before applying convolution neural networks, which significantly increases the computational complexity. 
To avoid the inefficient pre-computing process, the deconvolution layer and Standard Pixshuffle Layer (SPL), proposed by \cite{dong2016accelerating} and \cite{shi2016real} receptively, enable the networks to directly output HR images from LR images, which dramatically reduces the computational complexity, and contributes to recovery of more fine-grained details.

On the other hand, the deconvolution layer and SPL also make it possible for VSR networks \cite{yang2018drfn, wang2019edvr, yi2019progressive, tian2020tdan, isobe2020video, chen2020vesr} to output HR videos from LR ones directly.
The processing pipeline of the SOTA VSR methods is roughly as follows: extract features from the reference frame and neighboring frames, then feed them (after proper alignment and fusion) into a reconstruction network to generate a super-resolved frame.
By employing a deconvolution layer or an SPL in the reconstruction network, the SOTA VSR methods have been shown to generate satisfactory results in terms of efficiency and effectiveness on various datasets.

Although the SOTA SISR and VSR methods have performed satisfactorily on many datasets, they lack flexibility in adjusting the resolution of the final output. This is due to the intrinsic limitations of the deconvolution layer and the SPL.
Recently, the meta-up-sample module proposed by \cite{hu2019metasr} enables up-sampling by an arbitrary factor using a single model.
Its refined version, known as the scale-aware up-sampling module \cite{wang2020learning}, can better address the resulting memory overhead issues, but the underlying mechanism remains the same.

Unlike \cite{hu2019metasr, wang2020learning}, in the method proposed herein, we will generalize the SPL to release it from the constraints on the  up-sampling factors. 
It will be shown that the new mechanism performs on par with, or slightly better than, SPL in terms of fixed scale up-sampling, and delivers better performance than that in \cite{hu2019metasr, wang2020learning} in terms of up-sampling by arbitrary factors.

\subsection{Space-Time Video Super-Resolution}
Distinct from the separated operations of VFI and VSR, in a STVSR system we seek to simultaneously increase the temporal frame rate and the spatial resolution of a given video. This line of research was initiated in \cite{Shechtman2002}. As the STVSR operation is a highly ill-posed inverse problem, due to the inadequacy of the available information, traditional methods \cite{shechtman2005space, mudenagudi2010space, takeda2010spatiotemporal, shahar2011space} often resort to some hand-crafted priors or artificially-imposed constraints. 
For instance, reference \cite{shechtman2005space} adopts a space-time directional smoothness prior and reference \cite{mudenagudi2010space} makes a hypothesis that there is no great change in illumination for the static regions.
As a result, these methods cannot cope with many real-world scenarios. In addition, the optimization for these methods is extremely computational inefficient (e.g., the processing speed for \cite{mudenagudi2010space} is about $1$ min/frame).

With the aid of deep learning, it is now possible to develop data-driven assumption-free STVSR methods. One simple way to do that is to realize STVSR via sequential execution of deep-learning-based VFI and SR. However, this two-stage scheme is suboptimal since it is susceptible to error accumulation and makes no use of space-time correlation. In addition, a direct combination of VFI and SR without any consolidation is clearly inefficient in terms of the running cost.


In view of the problems with the two-stage approach, some one-stage STVSR methods \cite{haris2020space, xiang2020zooming} have been proposed, which are able to offer improved performance at a reduced cost. While they are highly innovative, these newly-proposed methods \cite{haris2020space, xiang2020zooming}, have two major limitations. Firstly, due to the use of a CNN to directly synthesize the intermediate frames, the temporal position of such frames is not adjustable after training. Secondly, there is no freedom to choose the spatial up-sampling factors to be different from those set in the training phase, nor to accommodate fractional factors.
The main motivation of the present work is to remove these two limitations and realize unconstrained STVSR.

After posting a preprint of this submission on arXiv \cite{shi2021learning}, we become aware of a concurrent and independent work \cite{xu2021temporal}, which addresses the temporal inflexibility problem using so-called TMBlock. But the spatial domain issue remains unsolved in \cite{xu2021temporal}. A performance comparison is included in Section \ref{sec:fix} of this submission.

\section{Unconstrained Space-Time Video Super-Resolution Network}
The goal of the proposed USTVSRNet is to transform a low-resolution low-frame-rate (LFR) video into a high-resolution high-frame-rate (HFR) one. Specifically, given two LR input frames ($I_0^{L}$ and $I_1^{L}$), an arbitrary target time $t\in[0, 1]$, and an arbitrary spatial up-sampling factor $s\in[1, +\infty)$, the goal is to synthesize an intermediate HR frame $I_t^{H}$ with $H=sL$. The overall architecture of USTVSRNet is shown in Fig.~\ref{fig:FrameWork}, which mainly consists of 4 sub-networks: a Frame Interpolation Network (FINet), a Feature Extractor, a Enhancement Network (EnhanceNet), and a Reconstruction Network.

As illustrated in Fig.~\ref{fig:FrameWork}, first a LR intermediate frame $\hat{I}_t^L$ is constructed by the FINet based on neighboring source frames ($I_0^{L}$ and $I_1^{L}$) and bidirectional optical flows (${f}_{t \rightarrow 0}$ and ${f}_{t \rightarrow 1}$). Then the features $F_0^L$, $F_t^L$ and $F_1^L$ are generated through the feature extractor from $I_0^L$, $\hat{I}_t^L$ and $I_1^L$ respectively. 
Next, $F_t^L$ is enhanced to $E_t^L$ at the feature level through the enhancement network, and, finally, $E_t^L$ is fed into the reconstruction network to produce a high-resolution frame $\hat{I}_t^H$ as an approximation of $I_t^H$. The details of each steps are outlined below.

\subsection{Frame Interpolation Network}
Given $I_0^L$ and $I_1^L$, the FINet is employed to generate a coarse prediction $\hat{I}_t^L$ as the reference frame, which will be used in conjunction with the feature-level prediction to produce the final reconstruction.
In principle, any flow-based VFI algorithm can serve this purpose. However, the SOTA systems \cite{jiang2018super, bao2019memc, bao2019depth, qvi_nips19} often involve complex designs (e.g., depth information \cite{bao2019depth}, quadratic model \cite{qvi_nips19}), and consequently are not very efficient as a component of a larger system. 
For this reason, we consider a simple design for the FINet.

First a light-weight optical flow estimation network (PWCNet \cite{sun2018pwc}) is utilized to estimate the bidirectional flows $f_{0 \rightarrow 1}$ and $f_{1 \rightarrow 0}$. They are then passed to the flow reverse layer \cite{qvi_nips19} to predict backward flows $f_{t \rightarrow 0}$ and $f_{t \rightarrow 1}$. Specifically, we have
\begin{equation}
f_{t \rightarrow 0} = \operatorname{FR}(f_{0 \rightarrow t}), 
\end{equation}
where $f_{0 \rightarrow t}=t*f_{0 \rightarrow 1}$, and $\operatorname{FR}$ denotes the flow reverse operation \cite{qvi_nips19}; $f_{t \rightarrow 1}$ can be computed in a similar way. Finally the reference frame is synthesized as:
\begin{equation}
\hat{I}_t^L = \frac{(1-t) \cdot B \cdot g(I_0^L, f_{t \rightarrow 0}) + t \cdot (1-B) \cdot g(I_1^L, f_{t \rightarrow 1})}{(1-t) \cdot B + t \cdot (1-B)},
\end{equation}
where $B$ is a blending mask generated by a small CNN \cite{bao2019memc, qvi_nips19, chi2020all}, and $g(\cdot)$ denotes the warping function.

\subsection{Feature Extractor}

The frame features $F_0^L$, $F_t^L$ and $F_1^L$ are extracted from $I_0^L$, $\hat{I}_t^L$ and $I_1^L$, respectively, through a feature extractor, which is composed of a convolution layer and several residual blocks \cite{he2016deep}.

\subsection{Enhancement Network}
\begin{figure}[h]
	\centering
	\begin{minipage}[h]{\linewidth}
		\centering
		\includegraphics[width=\linewidth]{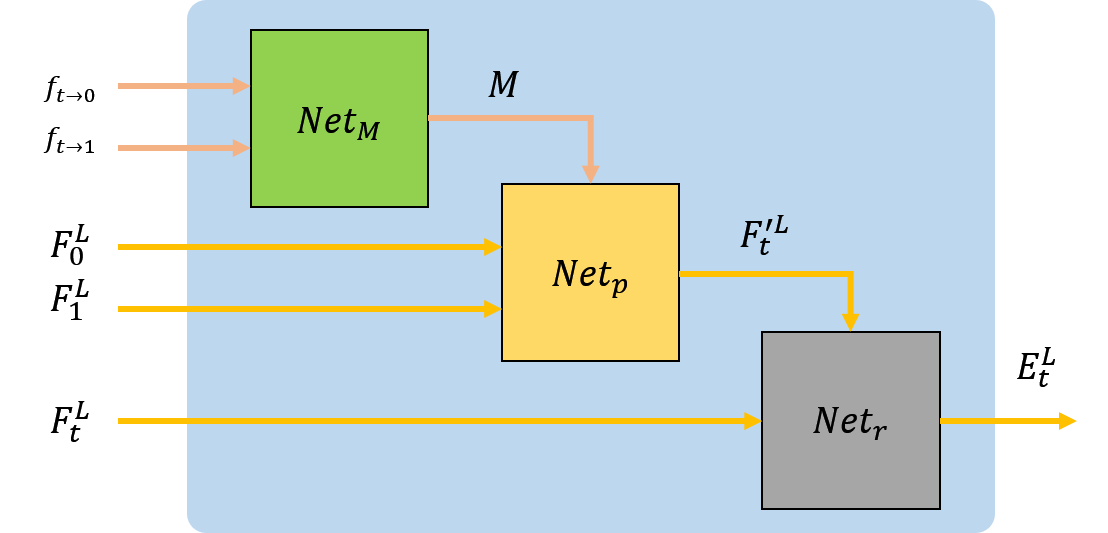}
		\centering
	\end{minipage}
	\figvspace
	\caption{Illustration of the architecture of EnhanceNet.}
	\label{fig:EnhanceNet}
\end{figure}

As illustrated in Fig.~\ref{fig:EnhanceNet}, the inputs to the enhancement network consist of the three extracted feature maps $F_0^L$, $F_t^L$, $F_1^L$ as well as the pre-computed bidirectional optical flows ${f}_{t \rightarrow 0}$, ${f}_{t \rightarrow 1}$. The goal of this sub-network is threefold: 
1) leverage the source frame features ($F_0^L$ and $F_1^L$) and the bidirectional optical flows (${f}_{t \rightarrow 0}$ and ${f}_{t \rightarrow 1}$) to predict the features of the intermediate frame $F_t^{'L}$; 2) refine the generated reference frame at the feature level to alleviate the error accumulation problem as the coarse prediction $\hat{I}_t^L$ obtained in the first stage tends to have many artifacts; 3) fuse the source frames to the intermediate frame for better reconstruction under the guidance of ${f}_{t \rightarrow 0}$, ${f}_{t \rightarrow 1}$.
The operation of the enhancement network can be expressed as:
\begin{equation}
M = \operatorname{Net}_M({f}_{t \rightarrow 0}, {f}_{t \rightarrow 1}),
\end{equation}
\begin{equation}
F_t^{'L} = \operatorname{Net}_p(F_0^L, F_1^L, M),
\end{equation}
\begin{equation}
E_t^L = \operatorname{Net}_r(F_t^L, F_t^{'L}),
\end{equation}
where $\operatorname{Net}_M$, $\operatorname{Net}_p$, and $\operatorname{Net}_r$ are composed of several residual blocks and a convolution layer,
and $M$ denotes the motion features extracted from ${f}_{t \rightarrow 0}$ and ${f}_{t \rightarrow 1}$ through $\operatorname{Net}_M$.

\subsection{Reconstruction Network}
The reconstruction network is designed using the residual dense network \cite{zhang2018residual} as the backbone. We replace the SPL with a novel GPL described below, making it possible to up-sample low-resolution features by an arbitrary scale factor $s$. Moreover, we substitute one out of every $K$ RDBs with 
our newly constructed SARDB, which is able to generate scale-adaptive features and contribute positively to the overall performance.

\subsubsection{Generalized Pixelshuffle Layer}
a new GPL is proposed to address the lack of flexibility in the SPL. Here we describe both the SPL and the GPL in parallel and highlight their differences.

The goal of the SPL and the GPL is to convert input feature maps of size $C_\text{in} \times H \times W$ to output feature maps of size $C_\text{out} \times sH \times sW$ for some scale factor $s$ ($s$ is allowed to be fractional for GPL but not for SPL). They both proceed in three steps:

\begin{figure}[t]
	\centering
	\begin{minipage}[h]{\linewidth}
		\centering
		\includegraphics[width=\linewidth]{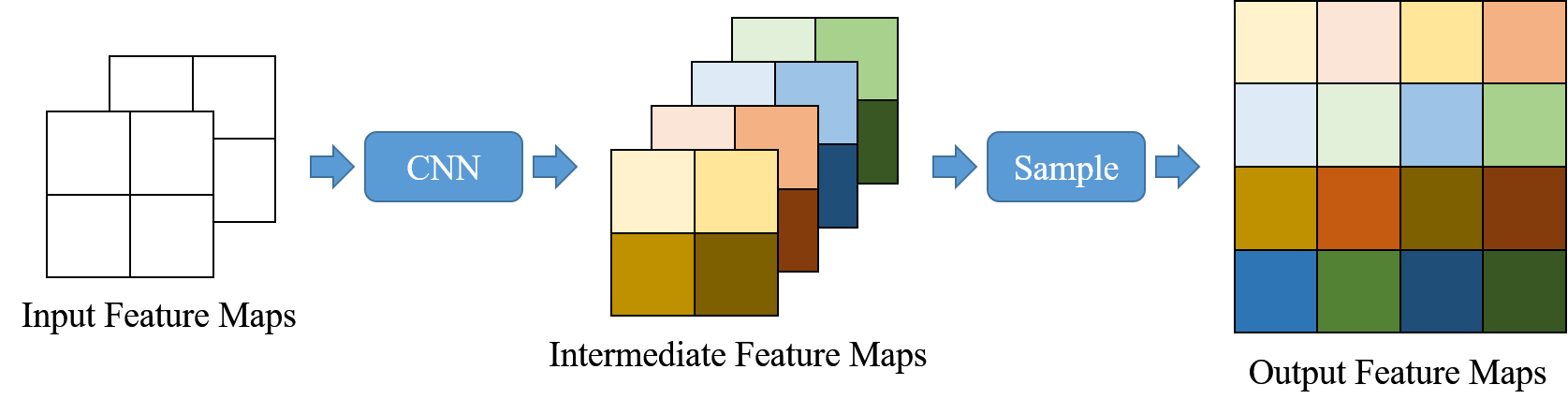}
		\scriptsize{(a)}
		\centering
	\end{minipage}
	\\
	\vspace{3mm}
	\begin{minipage}[h]{\linewidth}
		\centering
		\includegraphics[width=\linewidth]{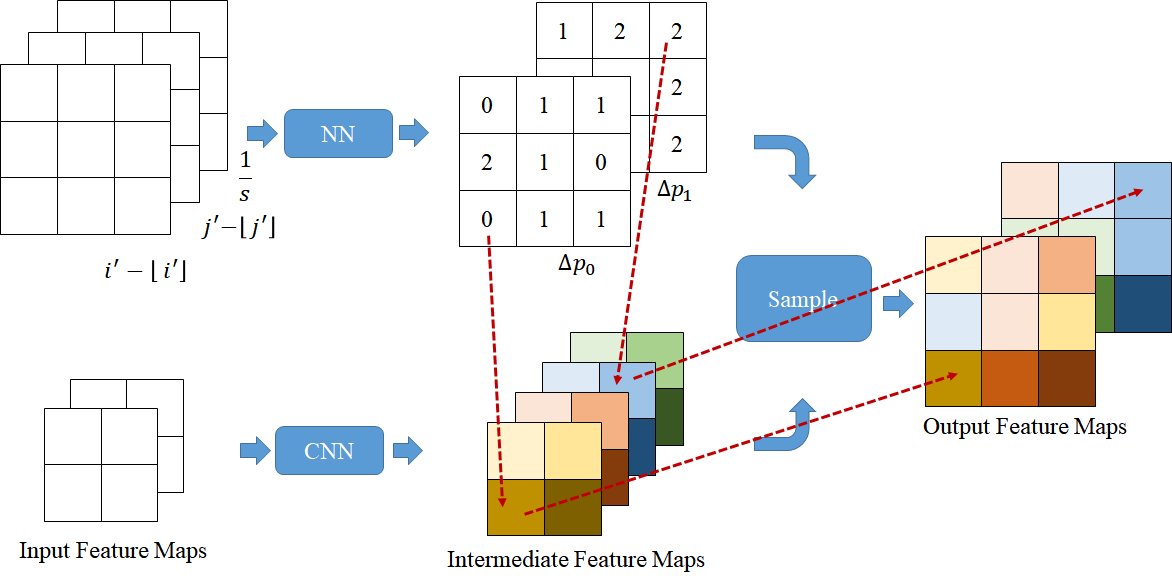}
		\scriptsize{(b)}
		\centering
	\end{minipage}
	\figvspace
	\caption{Examples of the standard and generalized pixelshuffle layers, where (a) shows the standard layer while $C_\text{in}=2, C_\text{int}=4, C_\text{out}=1$, and $s=2$; (b) shows the generalized pixshuffle layer while $C_\text{in}=2, C_\text{int}=4, C_\text{out}=2$, $s=1.5$, and $p_c=0$.}
	\label{fig:PL}
\end{figure}

\noindent \textbf{Widen Input Features:}
The input feature maps are transformed via convolution to the intermediate feature maps $T$ of size $C_\text{mid} \times H \times W$. Note that $C_\text{mid}$ must be equal to $s^2 C_\text{out}$ for SPL, but can be an arbitrary positive integer for GPL.

\noindent \textbf{Location Projection:}
Each spatial position on the output feature maps  $(i, j)$, $i \in [0, sH-1], j \in [0, sW-1]$ is projected to $({i}', {j}')=(\frac{i}{s}, \frac{j}{s})$ on the intermediate feature maps.

\noindent \textbf{Feature Mapping:}
Sample features from the intermediate feature maps $T$ for each 3D output position $(i, j, c), c \in [0, C_\text{out}-1]$ on the output feature maps according to a certain rule. Specifically, for SPL, the rule can be formulated as follows according to \cite{shi2016real}:
\begin{equation}
\operatorname{SPL}(T)_{i,j,c} = T_{\lfloor{i}'\rfloor, \lfloor{j}'\rfloor, C_\text{out}\cdot s \cdot \operatorname{mod}(i, s)+C_\text{out}\cdot s \cdot \operatorname{mod}(j, s)+c}.
\label{eq:spl}
\end{equation}
A concrete example can be found in Fig.~\ref{fig:PL} (a). In contrast, for GPL, we propose to sample using
\begin{equation}
\operatorname{GPL}(T)_{i,j,c} = T_{\lfloor{i}'\rfloor, \lfloor{j}'\rfloor, p_c+\triangle p_c},
\label{eq:gpl}
\end{equation}
where $p_c$ is a pre-determined channel position and $\triangle p_c$ denotes an adaptive offset predicted by a small fully connected network with $({i}' - \lfloor{i}'\rfloor, {j}'-\lfloor{j}'\rfloor, 1/s)$ as input (which is inspired by \cite{hu2019metasr}). {Note that we associate each 3D output position with a $\triangle p_c$, resulting in $sH \cdot sW \cdot C_\text{out}$ offsets in total}. In the case where $p_c+\triangle p_c$ is not an integer, the sampling value $T_{\lfloor{i}'\rfloor, \lfloor{j}'\rfloor, p_c+\triangle p_c}$ can be computed using a linear interpolation function:
\begin{equation}
T_{\lfloor{i}'\rfloor, \lfloor{j}'\rfloor, p_c+\triangle p_c} = \sum_{i=0}^{C_\text{out}-1}\max(0, 1-|p_c+\triangle p_c-i|) \cdot T_{\lfloor{i}'\rfloor, \lfloor{j}'\rfloor, i}.
\end{equation}
By designing so, the sampling position ($\lfloor{i}'\rfloor, \lfloor{j}'\rfloor, p_c+\triangle p_c$) on the intermediate feature maps is capable of moving along the channel direction to sample the needed feature. We provide a concrete example in Fig.~\ref{fig:PL} (b)

{From Eqs.~(\ref{eq:spl})-(\ref{eq:gpl}) and Fig.~\ref{fig:PL}, we have two observations:}
1) the proposed GPL not only achieves unconstrained up-sampling of feature maps but also has the capability to freely specify the channel dimension of the intermediate feature maps; 2) the GPL degenerates to the SPL if we set $C_\text{mid}=s^2C_\text{out}$, $p_c=C_\text{out}\cdot r \cdot \operatorname{mod}(i, s)+C_\text{out}\cdot r \cdot \operatorname{mod}(j, s)+c$, and force $\triangle p_c=0$. From these two points, it can be seen that the proposed GPL is a generalized version of the SPL with more degrees of freedom that can be fruitfully explored.

In our implementation, we set $p_c=c \cdot \frac{C_\text{mid}}{C_\text{out}} + \frac{C_\text{mid}/C_\text{out} - 1}{2}$. As such, the initial sampling positions are evenly distributed along the channel direction, which makes it possible to capture features as diverse as possible.
We initialize  $\triangle p_c$ with  $0$ and set the learning rate of the small fully connected layer  to be the same as the global learning rate.

\subsubsection{Scale-Attentive Residual Dense Block}
\begin{figure}[t]
	\centering
	\begin{minipage}[h]{\linewidth}
		\centering
		\includegraphics[width=\linewidth]{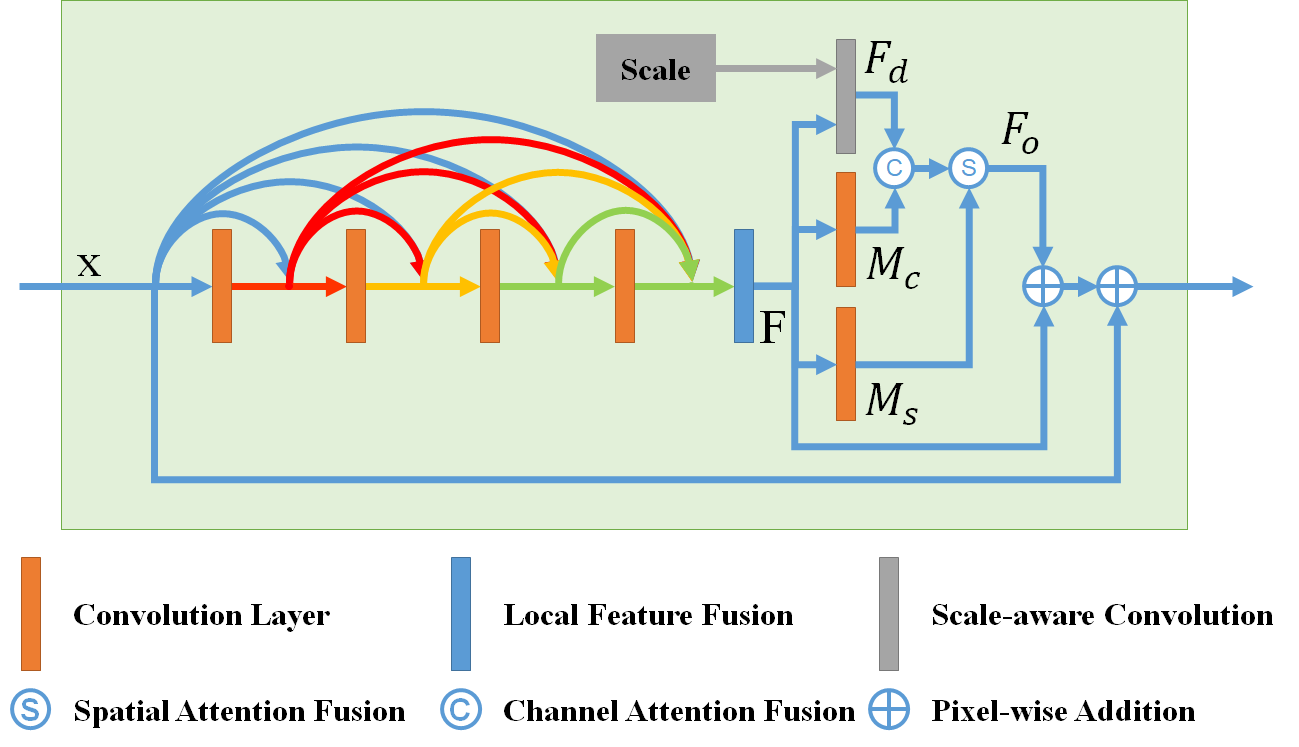}
		\centering
	\end{minipage}
	\figvspace
	\caption{Illustration of the SARDB architecture.}
	\label{fig:SARDB}
\end{figure}

As pointed out in \cite{wang2020learning}, the features generated by SR networks can be divided into scale-independent ones and scale-dependent ones, and the latter should be adapted to different scales. However, the scale-aware adaptation module introduced by \cite{wang2020learning} is built solely upon the spatial-wise attention mechanism, and makes no use of channel-wise attention \cite{woo2018cbam}. With this observation, we propose SARDB to exploit the available degrees of freedom more thoroughly.

The architecture of the proposed SARDB is shown in Fig.~\ref{fig:SARDB}.
The features $F$ output by the LFF \cite{zhang2018residual} are fed into several convolution layers to generate spatial attention map $M_s$ and channel attention map $M_c$ respectively. Then, the scale-aware convolution \cite{wang2020learning} is employed to convert the features $F$ into scale-dependent features $F_{d}$, which are then modulated by $M_s$ and $M_c$ by broadcasting and multiplication. The above operations can be expressed as follows:
\begin{equation}
\begin{aligned}
F_d &= \operatorname{Sconv}(F),\\
M_c &= \operatorname{Net_c}(F),\\
M_s &= \operatorname{Net_s}(F),\\
F_o &= F_d \odot M_c \odot M_s + F.
\end{aligned}
\end{equation}
Finally, the results from the upper branch $F_o$ and lower branches $x$ are merged to produce scale-adaptive features.

\section{Experiments For Unconstrained Space-Time Video Super-Resolution}
Unconstrained STVSR methods can flexibly adjust the temporal frame rate and the spatial resolution of the output video. In this section, we discuss the unconstrained STVSR. The experiment for fixed STVSR will be presented in Section \ref{sec:fix}.

\subsection{Implementation Details}
In our experiments, we explore the performance for different values of the target time $t$ and the up-sampling factor $s$. We let $t$ vary from $0$ to $1$ with a step size of $0.125$, and $s$ vary from $1$ to $4$ with a step of $0.5$. During the implementation, we set $K=4$ and $C_\text{mid} = 5C_\text{in} = 5C_\text{out}=5\times64$ respectively. The adopted loss function, training dataset, and training strategy are described below.

\subsubsection[noindent]{Loss Function}
We employ two loss terms to train our network, being $L_1 $ loss and perceptual loss \cite{johnson2016perceptual}, respectively:

\begin{figure*}[t]
	\centering
	\begin{minipage}[b]{0.30\linewidth}
		\centering
		\includegraphics[width=\linewidth]{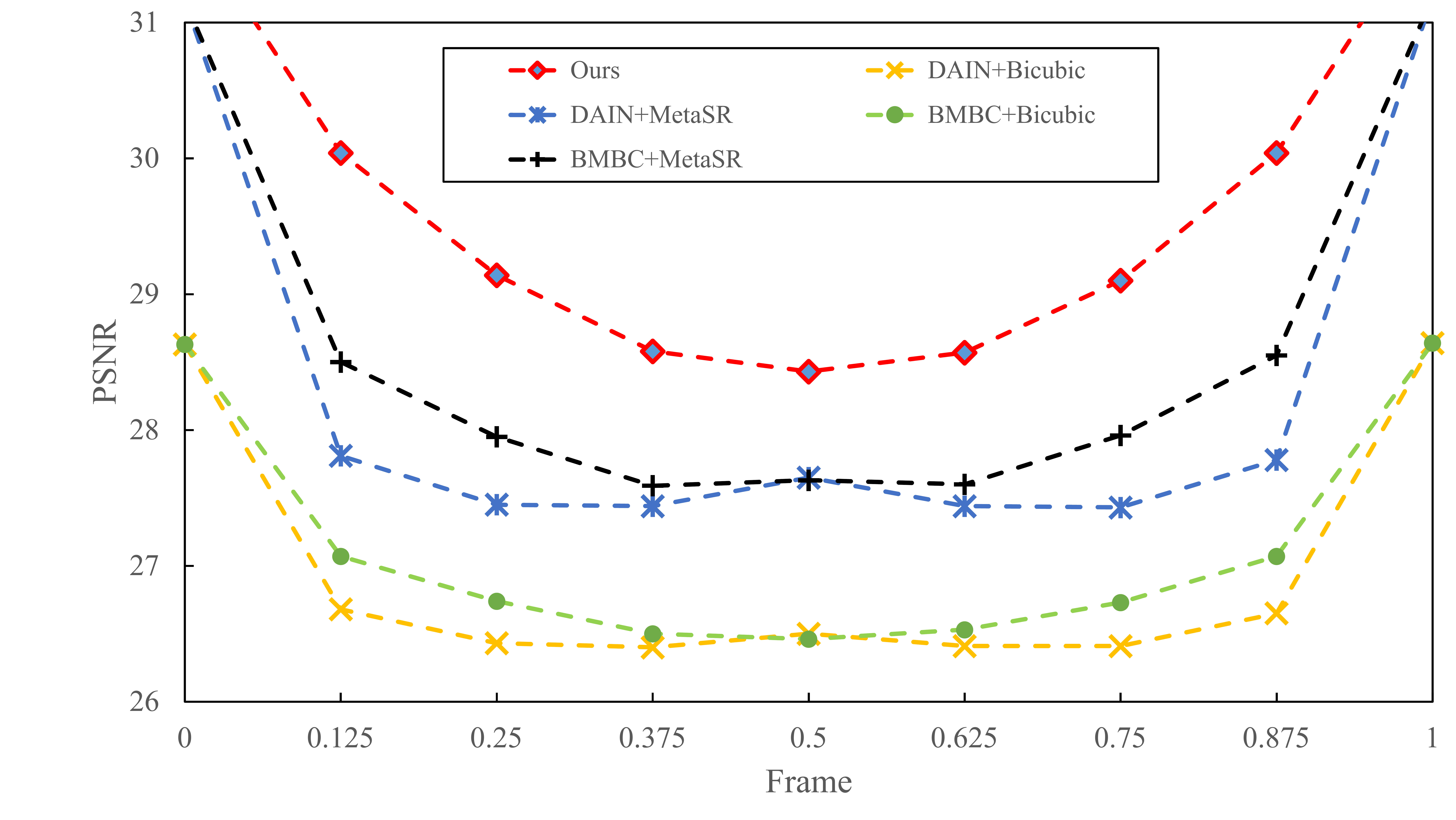}
		\scriptsize{(a) PSNR for different temporal positions with $s=2.5$}
	\end{minipage}
	\hspace{5mm}
	\begin{minipage}[b]{0.30\linewidth}
		\centering
		\includegraphics[width=\linewidth]{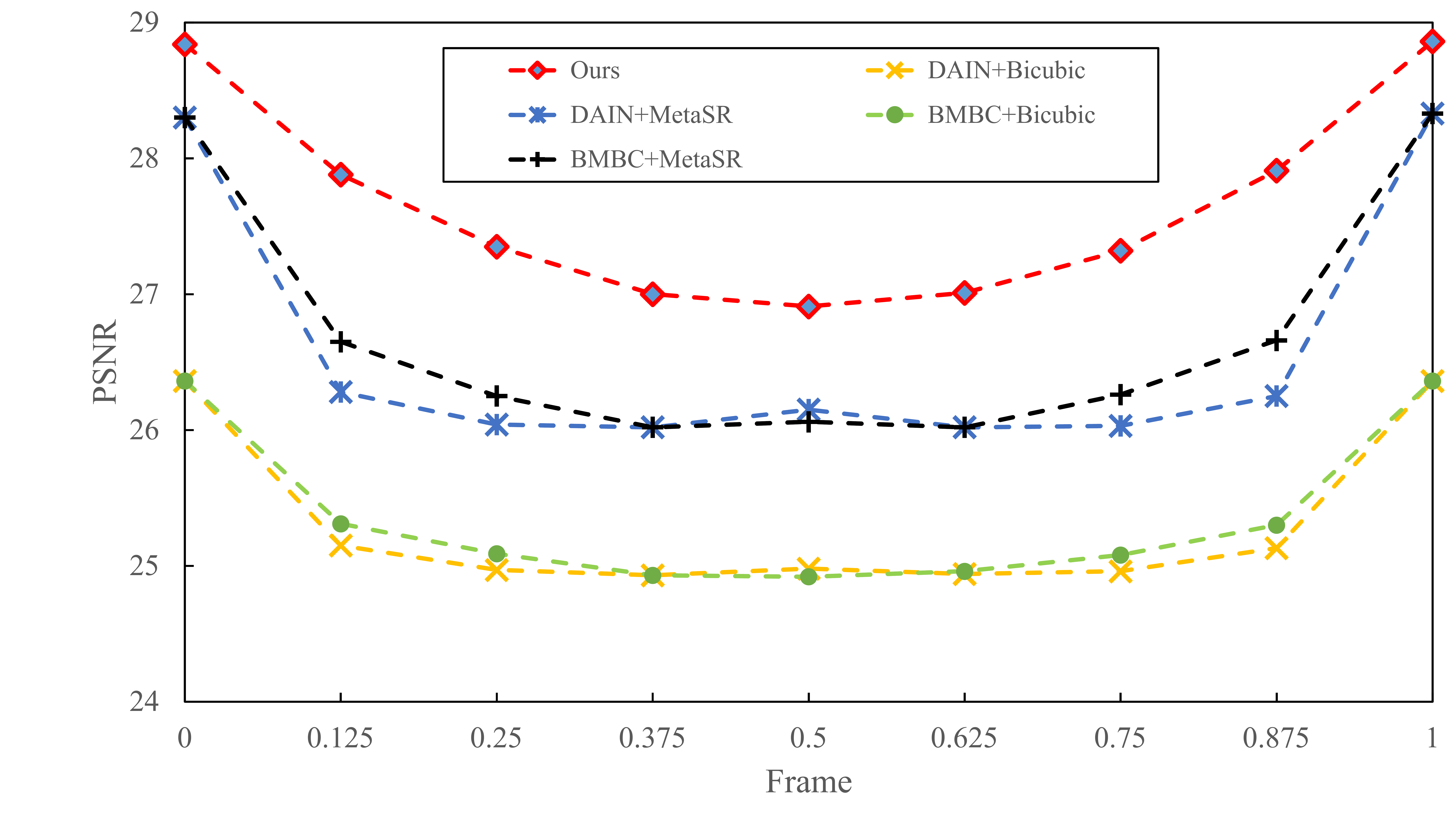}
		\scriptsize{(b) PSNR for different temporal positions with $s=3.5$}
	\end{minipage}
	\hspace{5mm}
	\begin{minipage}[b]{0.30\linewidth}
		\centering
		\includegraphics[width=\linewidth]{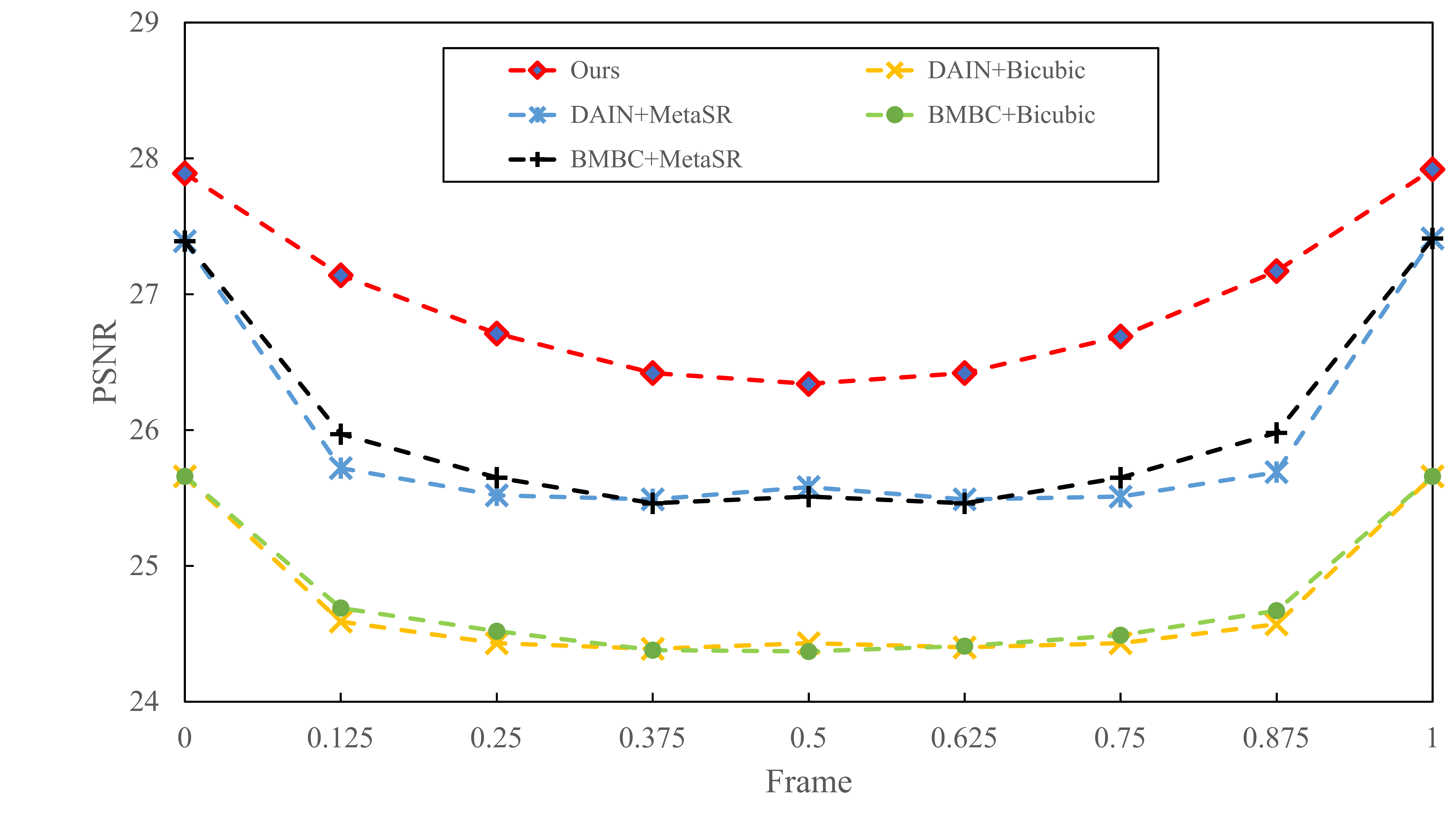}
		\scriptsize{(c) PSNR for different temporal positions with $s=4.0$}
	\end{minipage}

	\begin{minipage}[b]{0.30\linewidth}
		\centering
		\includegraphics[width=\linewidth]{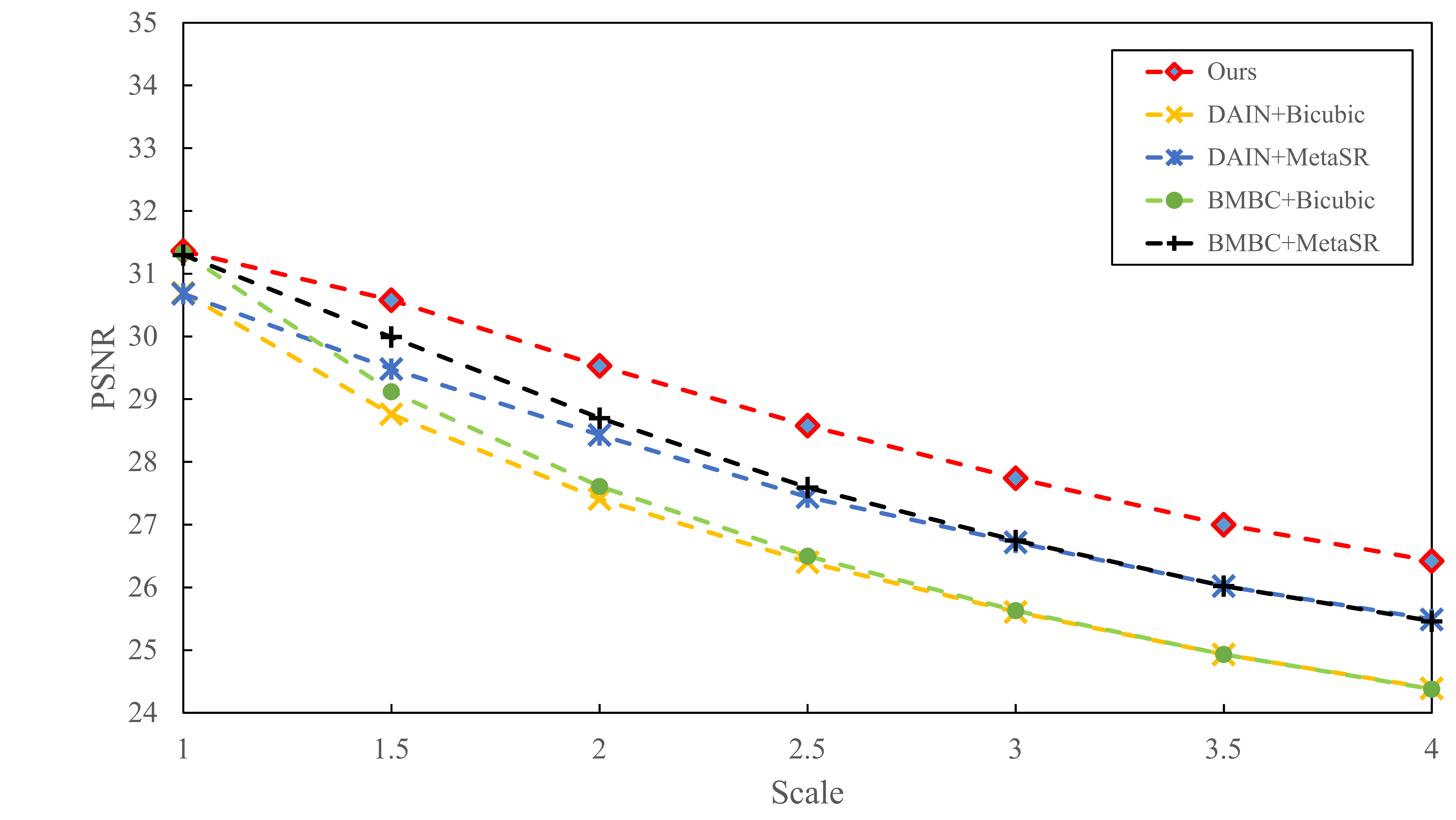}
		\scriptsize{(d) PSNR for different scale factors with $t=0.375$}
	\end{minipage}
	\hspace{5mm}
	\begin{minipage}[b]{0.30\linewidth}
		\centering
		\includegraphics[width=\linewidth]{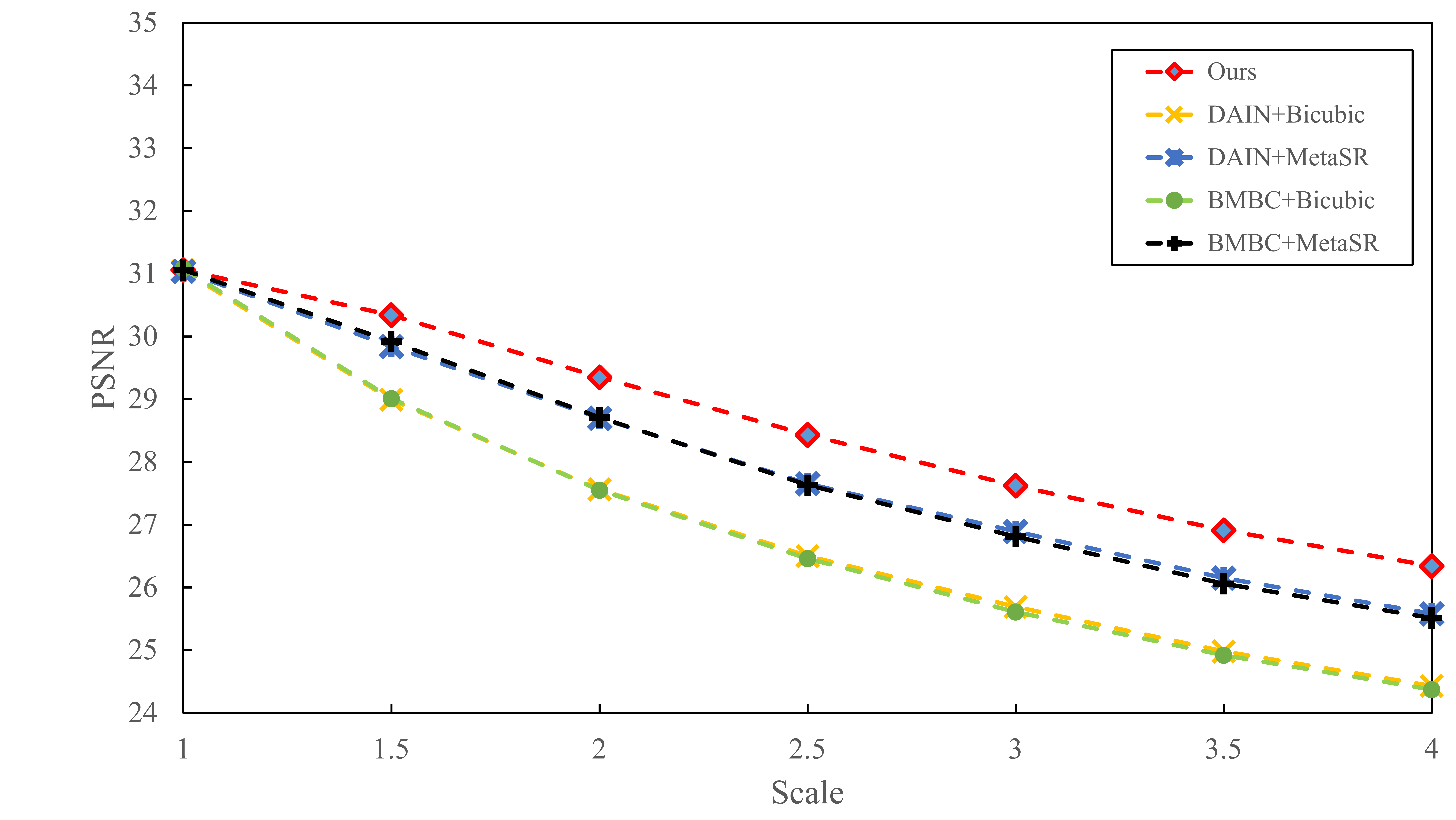}
		\scriptsize{(e) PSNR for different scale factors with $t=0.500$}
	\end{minipage}
	\hspace{5mm}
	\begin{minipage}[b]{0.30\linewidth}
		\centering
		\includegraphics[width=\linewidth]{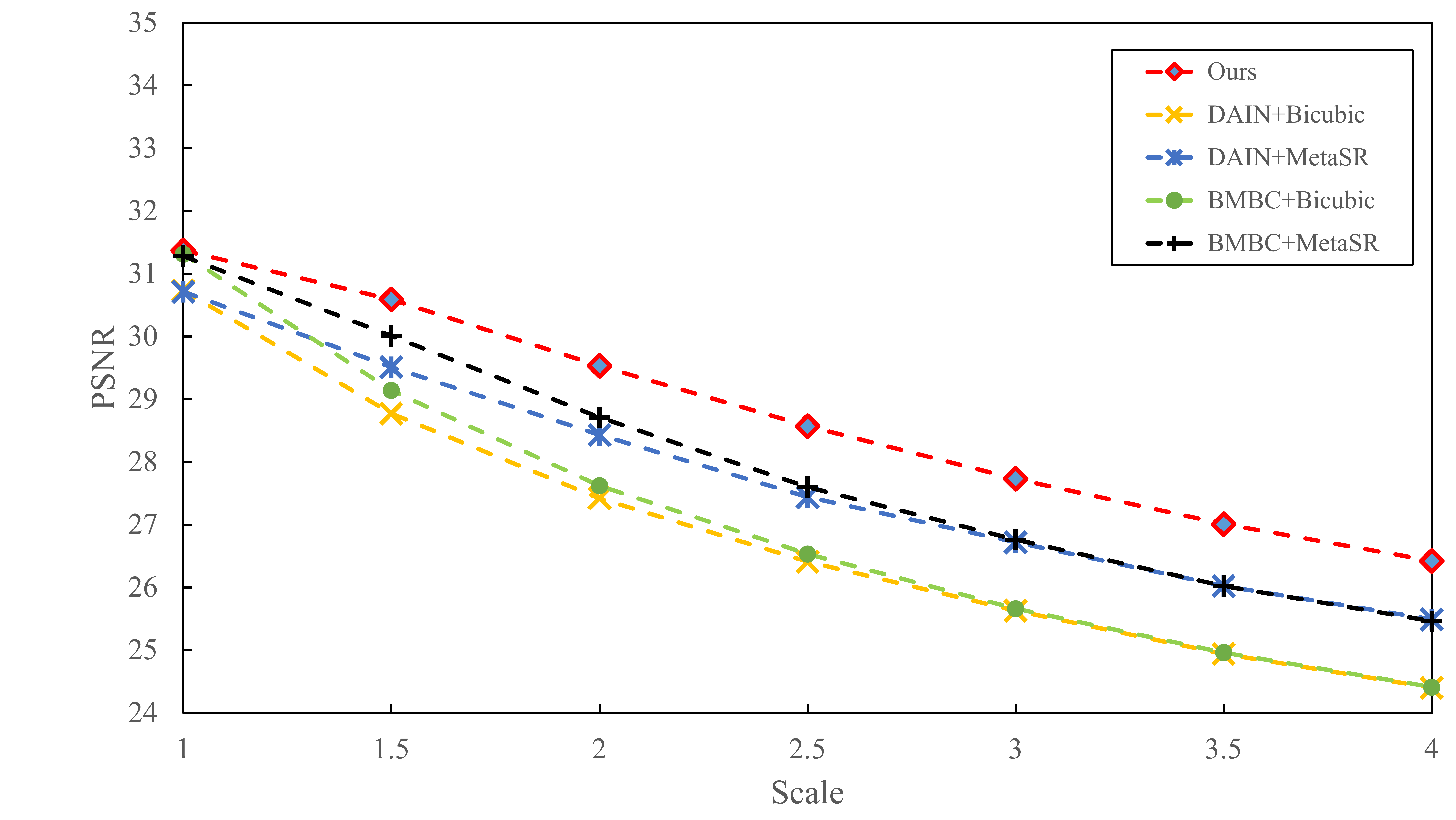}
		\scriptsize{(f) PSNR for different scale factors with $t=0.625$}
	\end{minipage}
	\caption{Quantitative comparisons of unconstrained STVSR methods on Adobe240 dataset.}
	\label{fig:quantitative comparsion adobe}
\end{figure*}

\noindent \textbf{Reconstruction Loss:}
The $L_1$ loss is used to measure the difference between the prediction and the ground-truth in a per-pixel manner, and can be formulated as follows:
\begin{equation}
\begin{aligned}
\mathcal{L}_1 = \sum_{x} ||\hat{I}_t^H(x) - I_t^H(x)||_1.
\end{aligned}
\end{equation}
The $L_2$ loss can also be used, but it is widely known in the image synthesis area that the $L_2$ loss could lead to blurry results to a certain degree.
Following \cite{lee2020adacof, liu2017video}, we adopt Charbonnier penalty function \cite{charbonnier1994two} to optimize $L_1$ loss function and set $\epsilon=10^{-6}$.

\begin{figure*}[t]
	\centering
	\begin{minipage}[b]{0.30\linewidth}
		\centering
		\includegraphics[width=\linewidth]{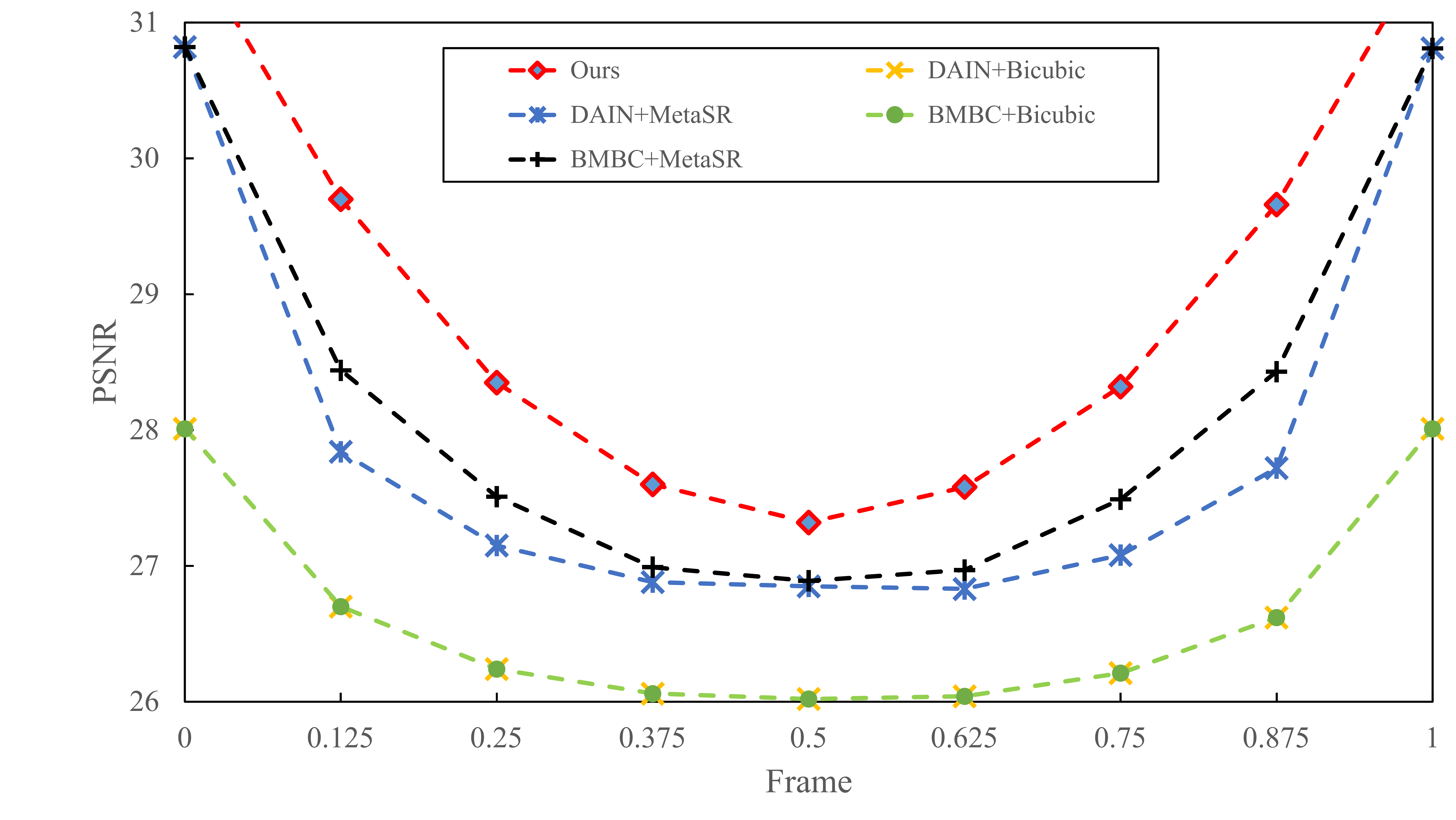}
		\scriptsize{(a) PSNR for different temporal positions with $s=2.5$}
	\end{minipage}
	\hspace{5mm}
	\begin{minipage}[b]{0.30\linewidth}
		\centering
		\includegraphics[width=\linewidth]{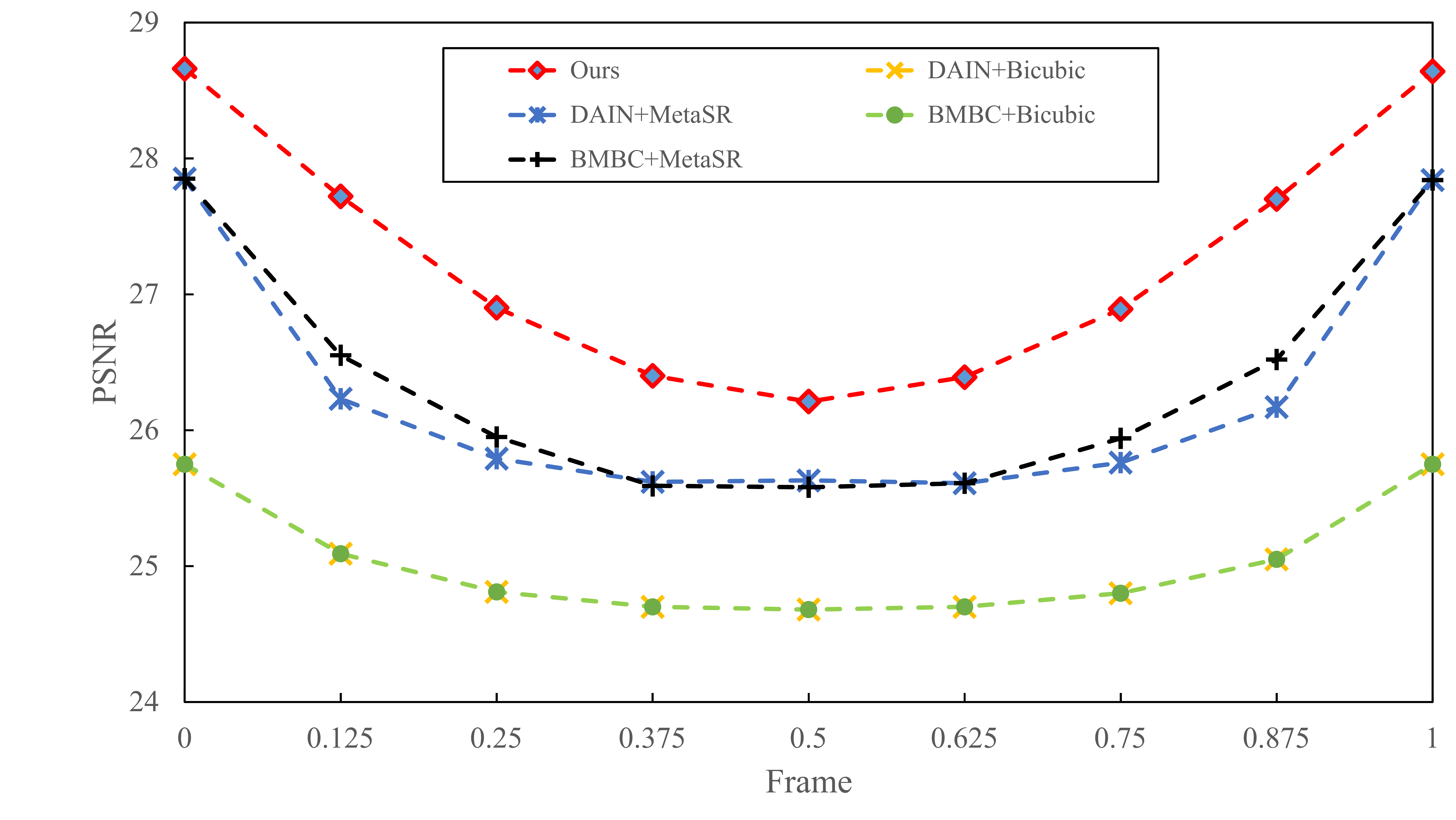}
		\scriptsize{(b) PSNR for different temporal positions with $s=3.5$}
	\end{minipage}
	\hspace{5mm}
	\begin{minipage}[b]{0.30\linewidth}
		\centering
		\includegraphics[width=\linewidth]{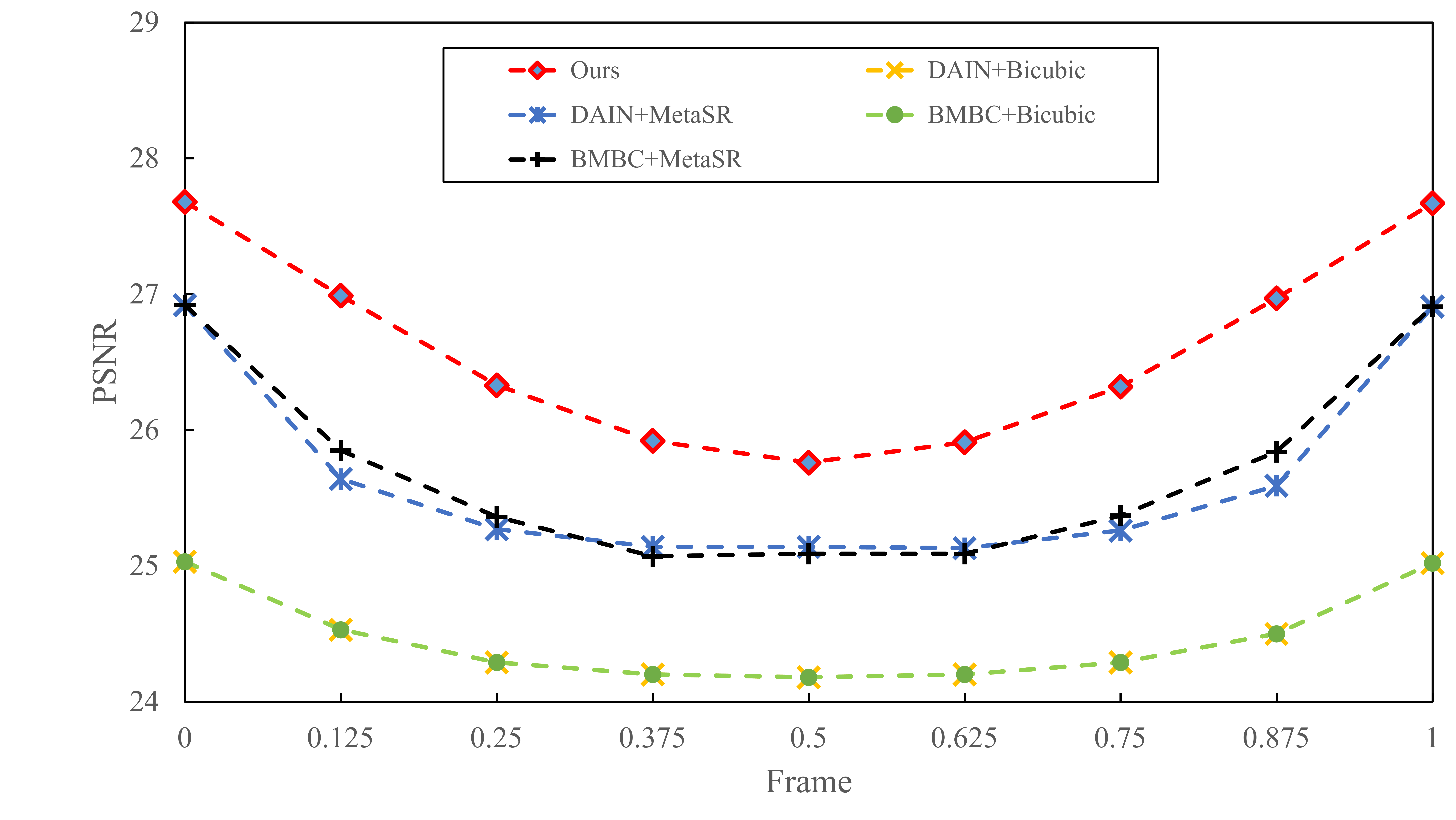}
		\scriptsize{(c) PSNR for different temporal positions with $s=4.0$}
	\end{minipage}
	\begin{minipage}[b]{0.30\linewidth}
		\centering
		\includegraphics[width=\linewidth]{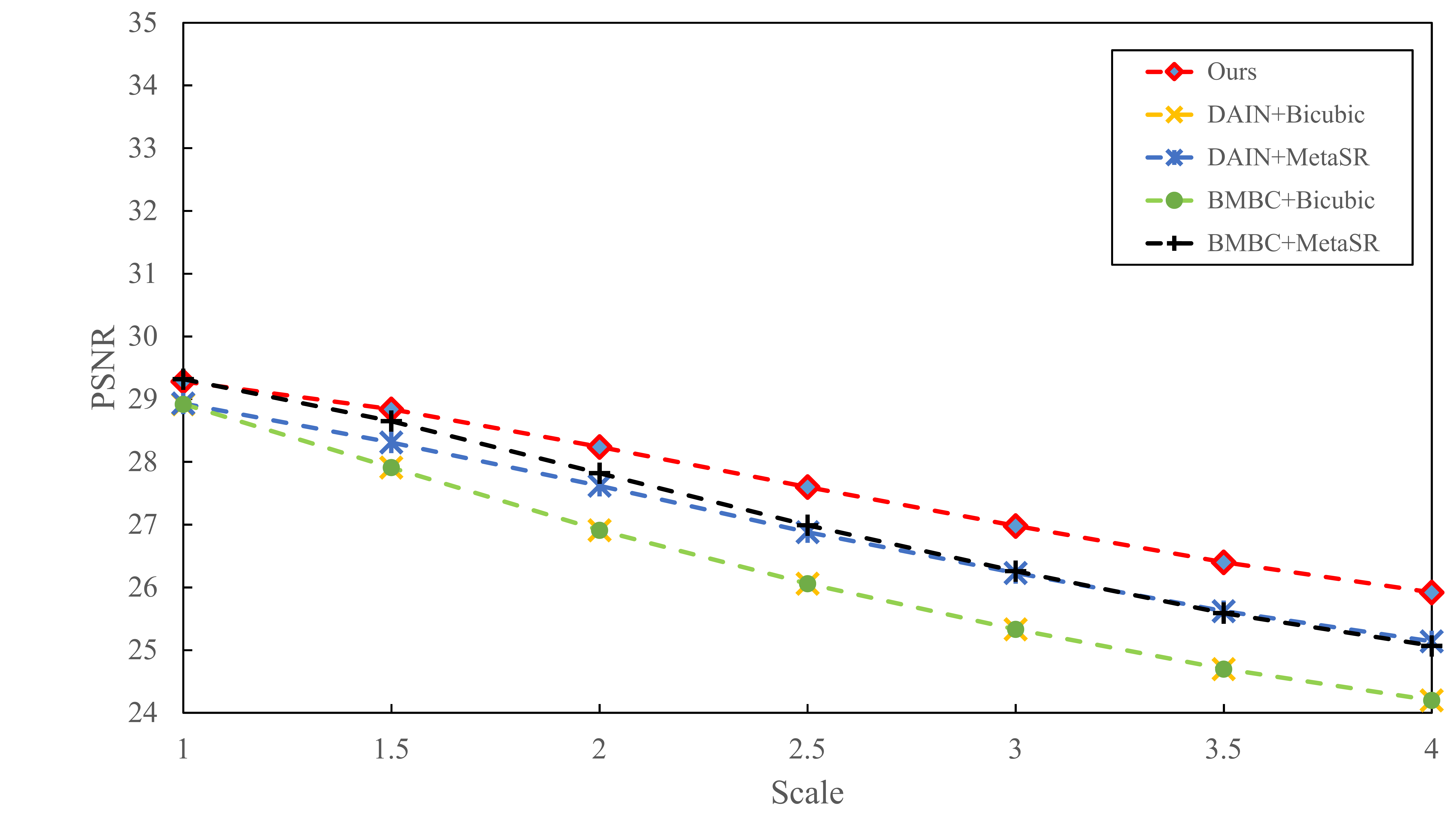}
		\scriptsize{(d) PSNR for different scale factors with $t=0.375$}
	\end{minipage}
	\hspace{5mm}
	\begin{minipage}[b]{0.30\linewidth}
		\centering
		\includegraphics[width=\linewidth]{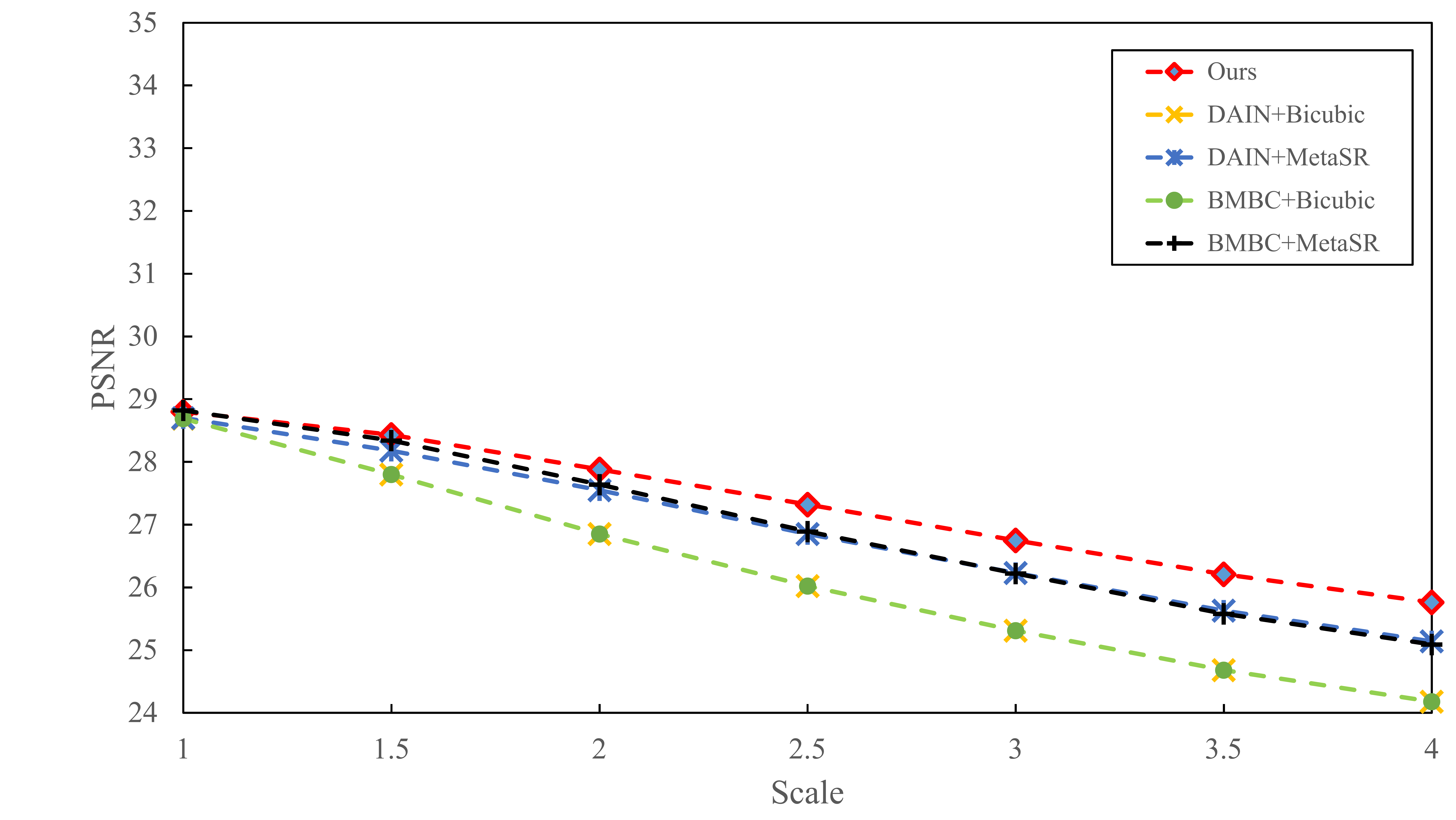}
		\scriptsize{(e) PSNR for different scale factors with $t=0.500$}
	\end{minipage}
	\hspace{5mm}
	\begin{minipage}[b]{0.30\linewidth}
		\centering
		\includegraphics[width=\linewidth]{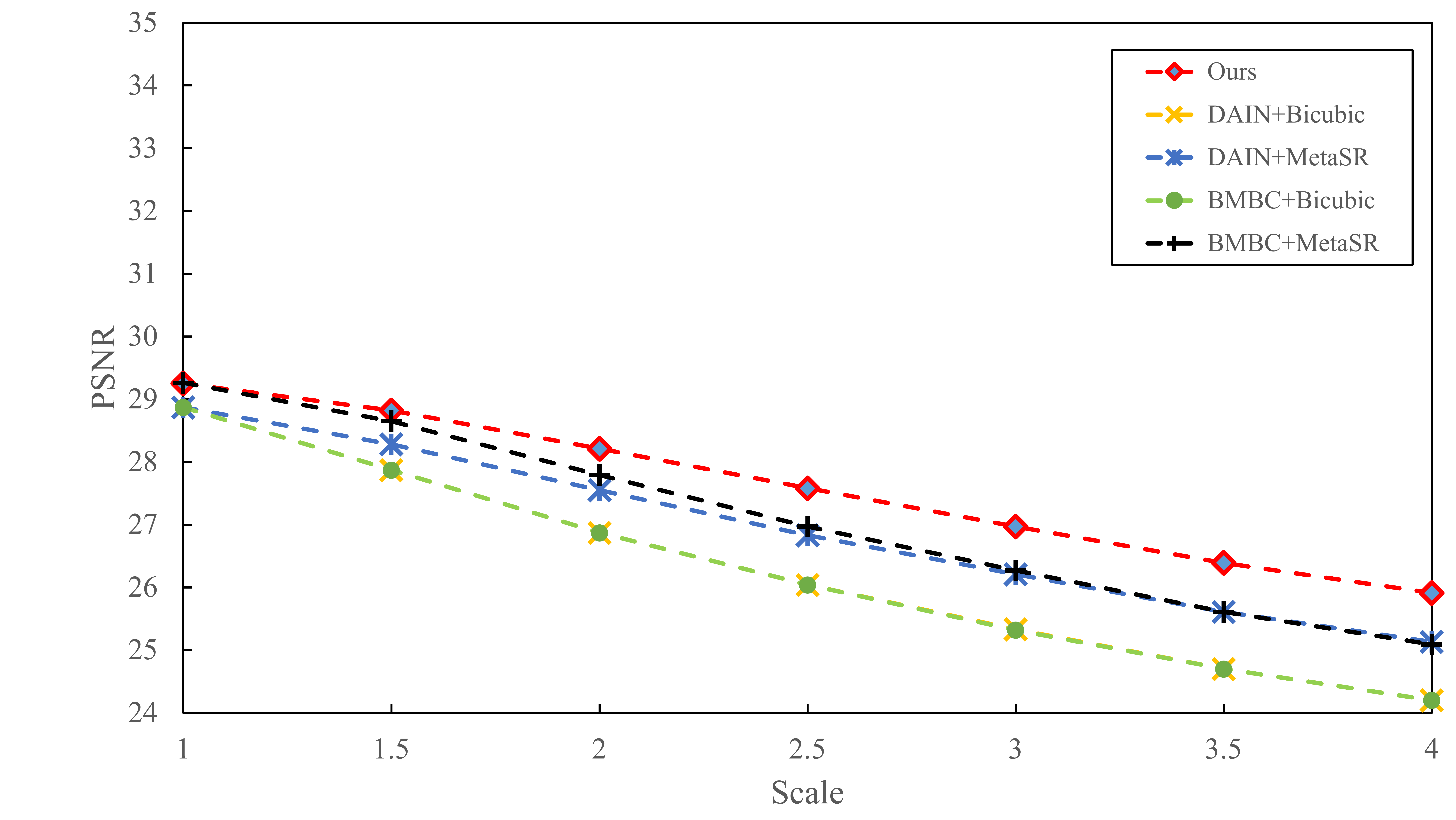}
		\scriptsize{(f) PSNR for different scale factors with $t=0.625$}
	\end{minipage}
	\caption{Quantitative comparisons of unconstrained STVSR methods on Gopro dataset.}
	\label{fig:quantitative comparsion gopro}
\end{figure*}

\begin{figure*}[t]
	\begin{center}
		\begin{minipage}[htb]{0.16\linewidth}
			\centering
			{\includegraphics[width=\linewidth]{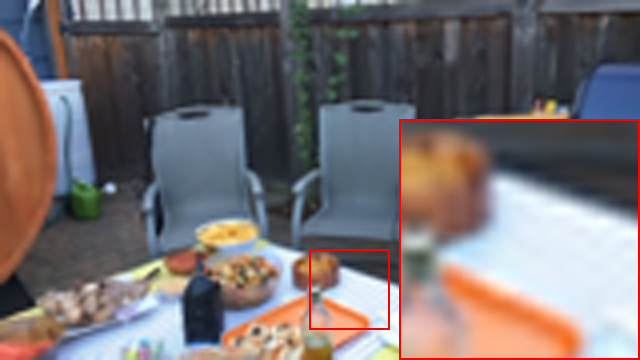}}\\
			\intervspace
			{\includegraphics[width=\linewidth]{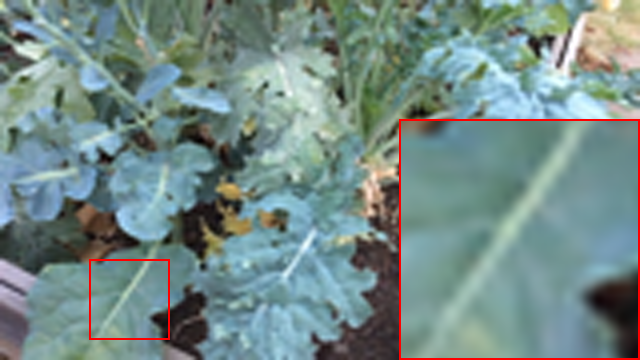}}\\
			\intervspace
			{\includegraphics[width=\linewidth]{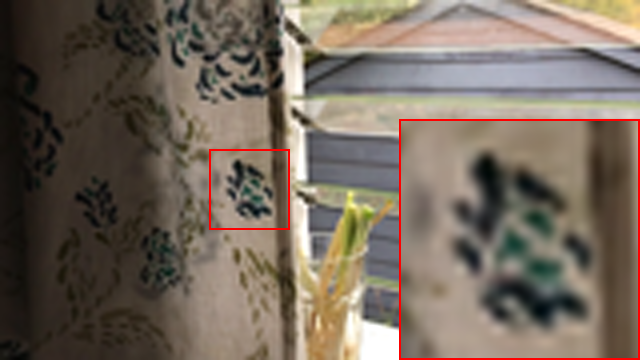}}\\
			\intervspace
			{\includegraphics[width=\linewidth]{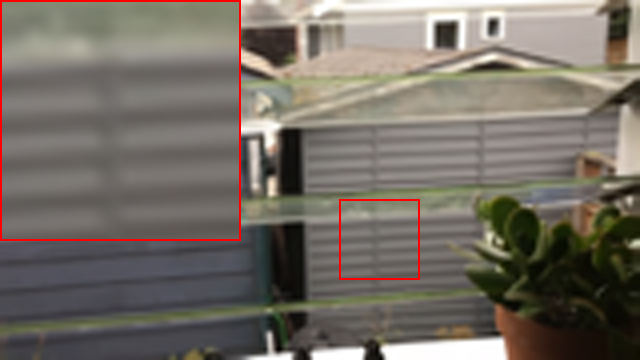}}\\
			\intervspace
			{\includegraphics[width=\linewidth]{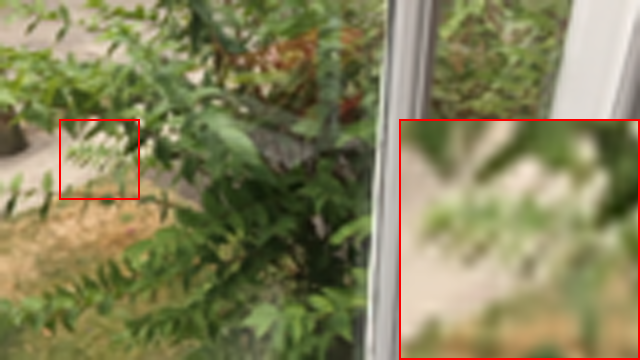}}\\
			\intervspace
			{\includegraphics[width=\linewidth]{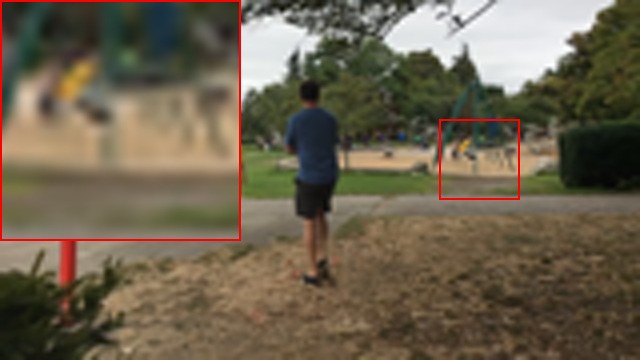}}\\
			\intervspace
			{\includegraphics[width=\linewidth]{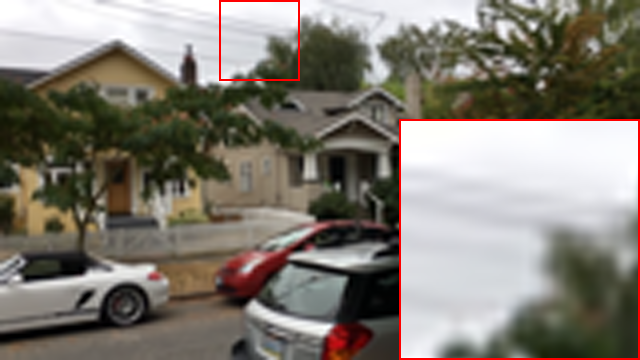}}\\
			\intervspace
			{\includegraphics[width=\linewidth]{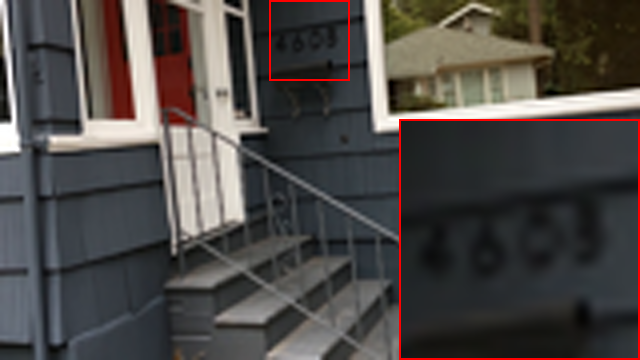}}\\
			\intervspace
			\intervspace
			\centerline{DAIN+Bicubic}
		\end{minipage}
		\begin{minipage}[htb]{0.16\linewidth}
			\centering
			{\includegraphics[width=\linewidth]{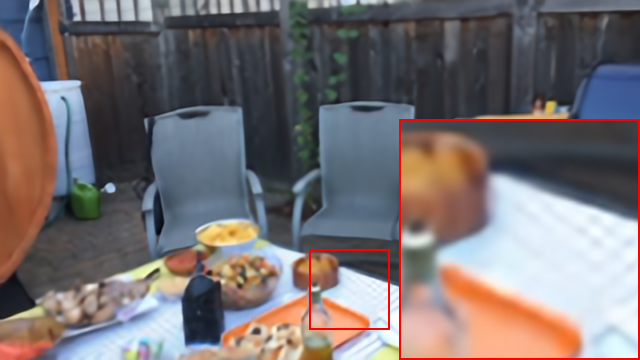}}\\
			\intervspace
			{\includegraphics[width=\linewidth]{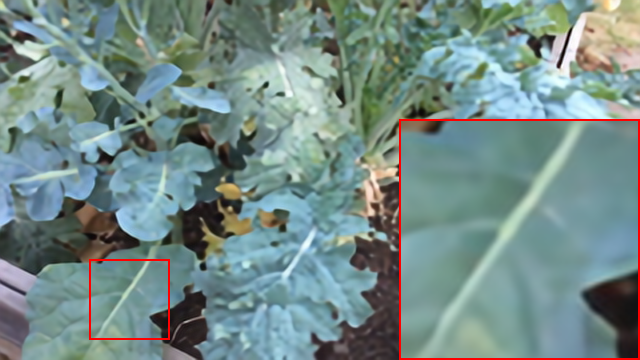}}\\
			\intervspace
			{\includegraphics[width=\linewidth]{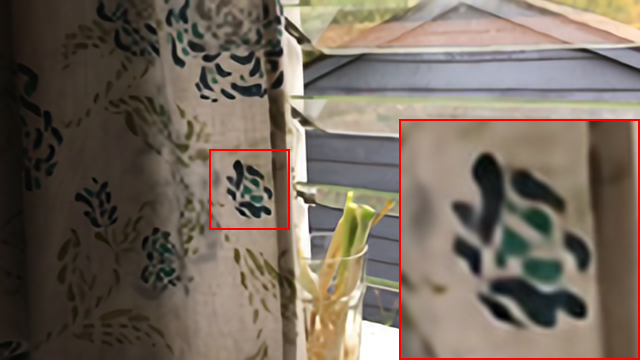}}\\
			\intervspace
			{\includegraphics[width=\linewidth]{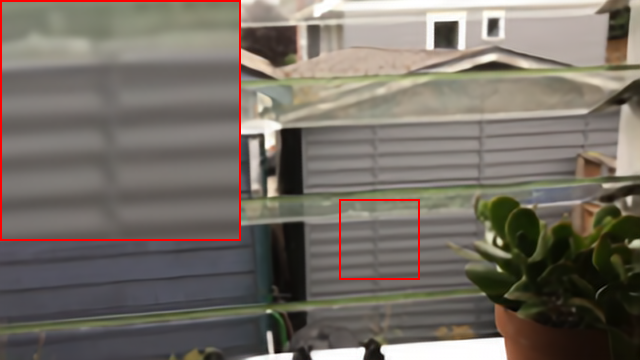}}\\
			\intervspace
			{\includegraphics[width=\linewidth]{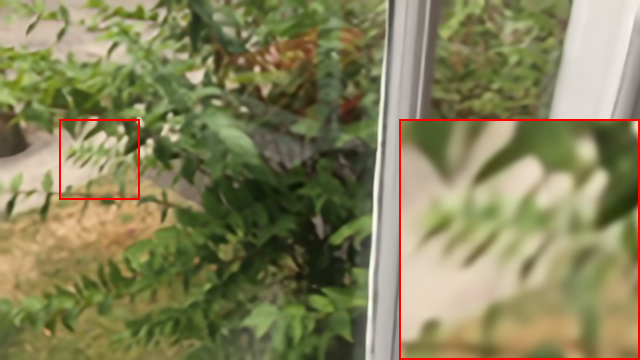}}\\
			\intervspace
			{\includegraphics[width=\linewidth]{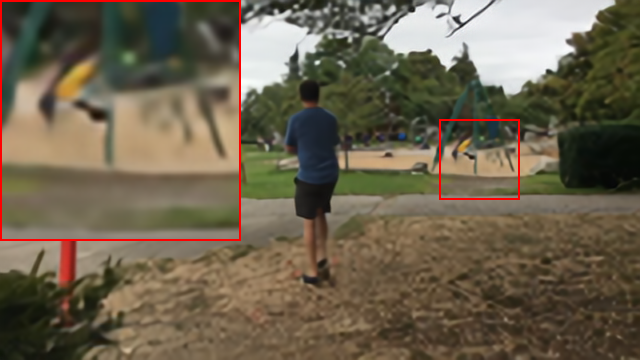}}\\
			\intervspace
			{\includegraphics[width=\linewidth]{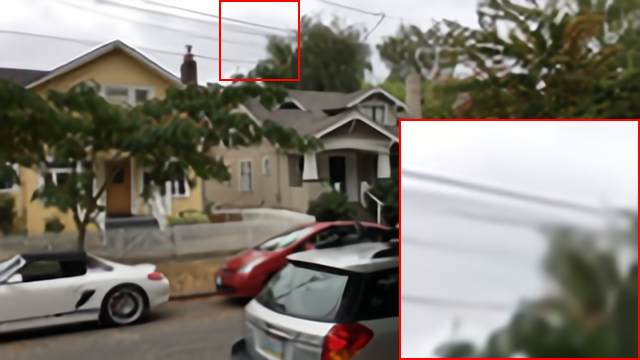}}\\
			\intervspace
			{\includegraphics[width=\linewidth]{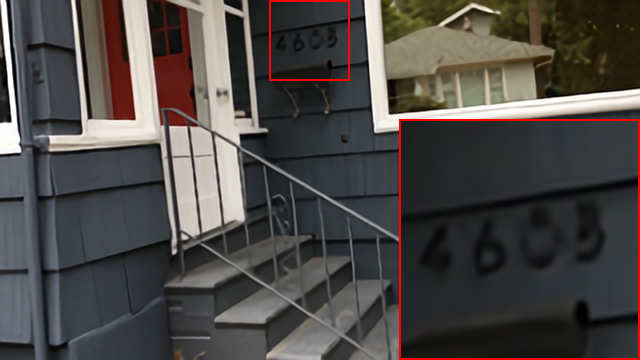}}\\
			\intervspace
			\intervspace
			\centerline{DAIN+MetaSR}
		\end{minipage}
		\begin{minipage}[htb]{0.16\linewidth}
			\centering
			{\includegraphics[width=\linewidth]{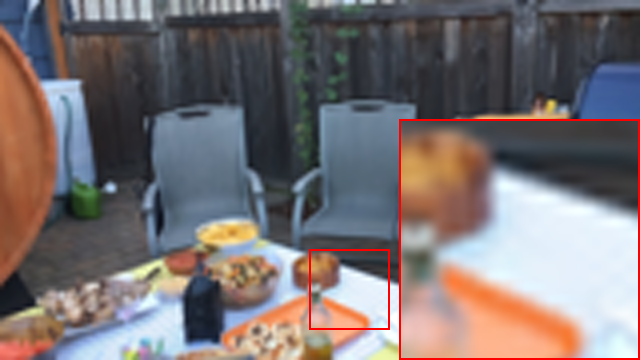}}\\
			\intervspace
			{\includegraphics[width=\linewidth]{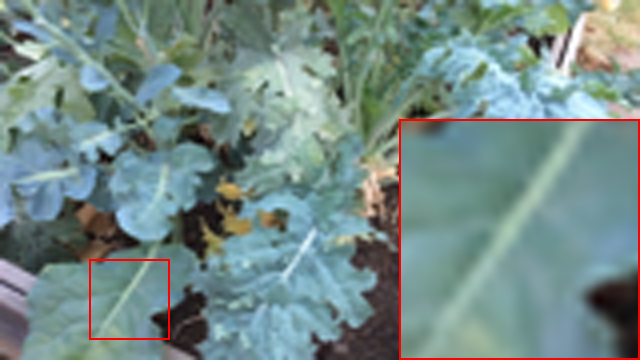}}\\
			\intervspace
			{\includegraphics[width=\linewidth]{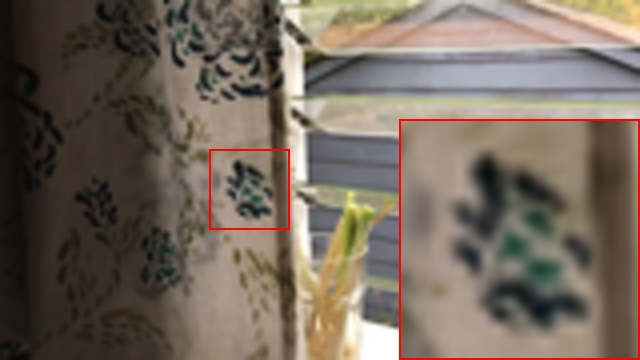}}\\
			\intervspace
			{\includegraphics[width=\linewidth]{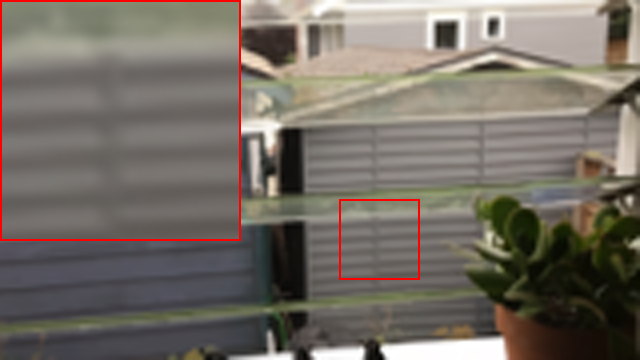}}\\
			\intervspace
			{\includegraphics[width=\linewidth]{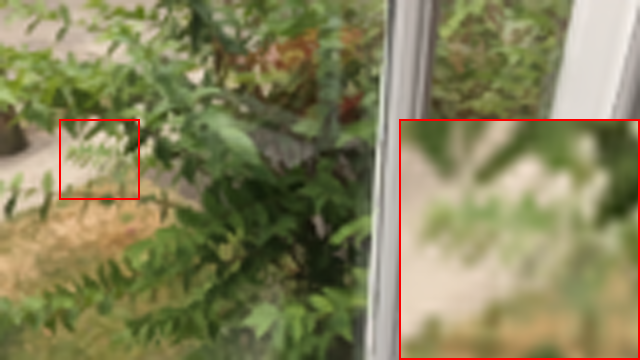}}\\
			\intervspace
			{\includegraphics[width=\linewidth]{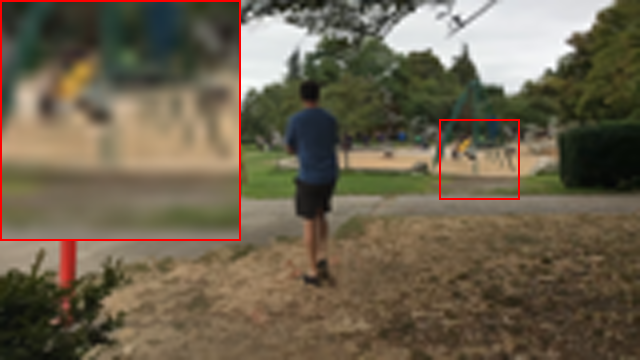}}\\
			\intervspace
			{\includegraphics[width=\linewidth]{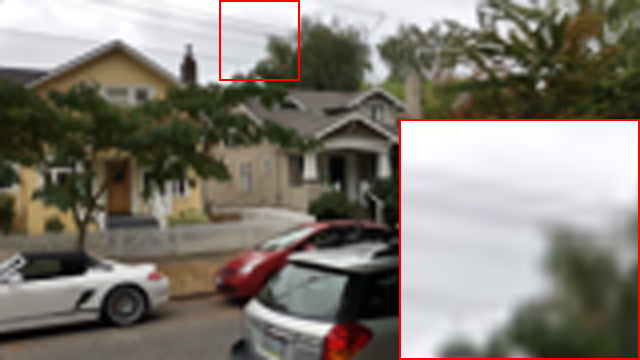}}\\
			\intervspace
			{\includegraphics[width=\linewidth]{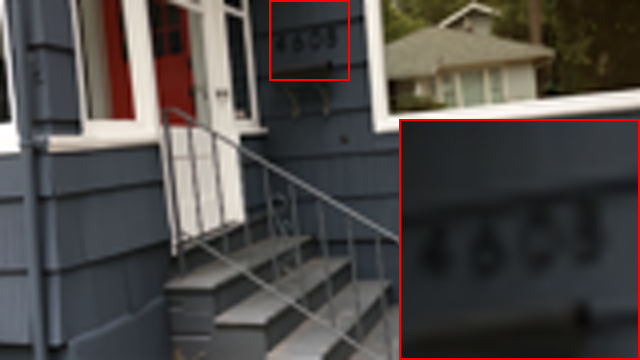}}\\
			\intervspace
			\intervspace
			\centerline{BMBC+Bicubic}
		\end{minipage}
		\begin{minipage}[htb]{0.16\linewidth}
			\centering
			{\includegraphics[width=\linewidth]{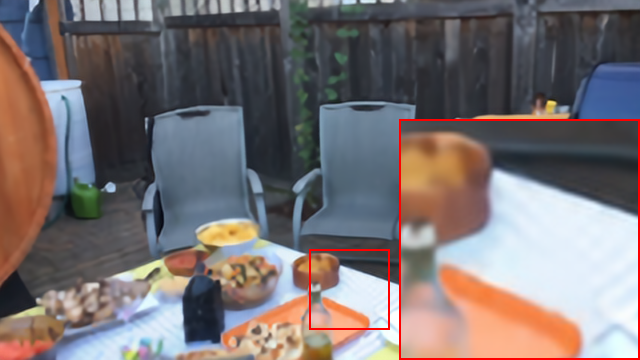}}\\
			\intervspace
			{\includegraphics[width=\linewidth]{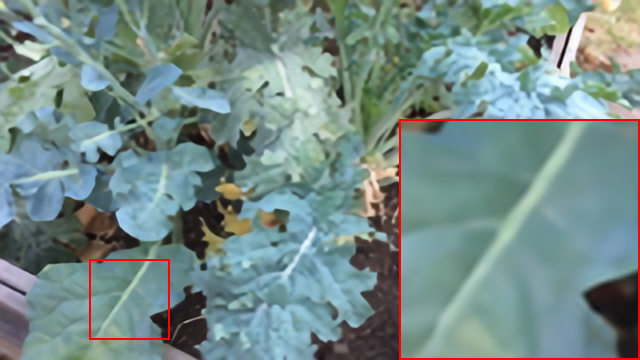}}\\
			\intervspace
			{\includegraphics[width=\linewidth]{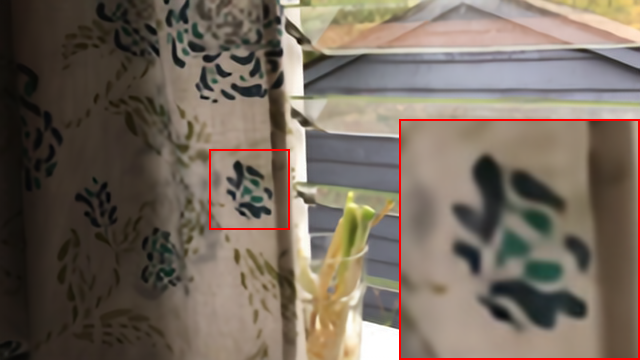}}\\
			\intervspace
			{\includegraphics[width=\linewidth]{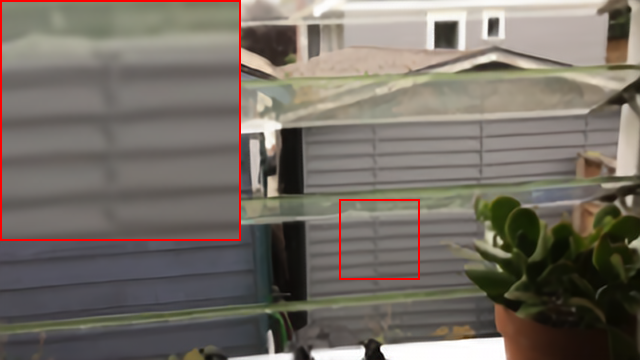}}\\
			\intervspace
			{\includegraphics[width=\linewidth]{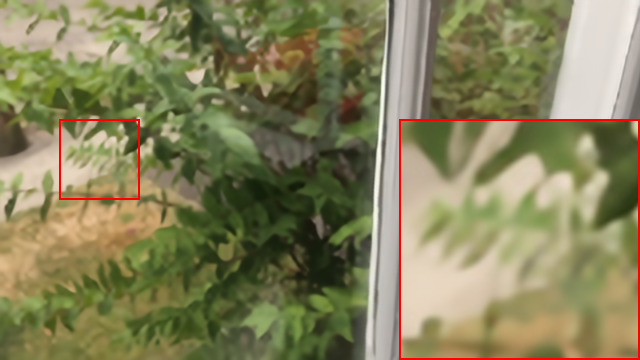}}\\
			\intervspace
			{\includegraphics[width=\linewidth]{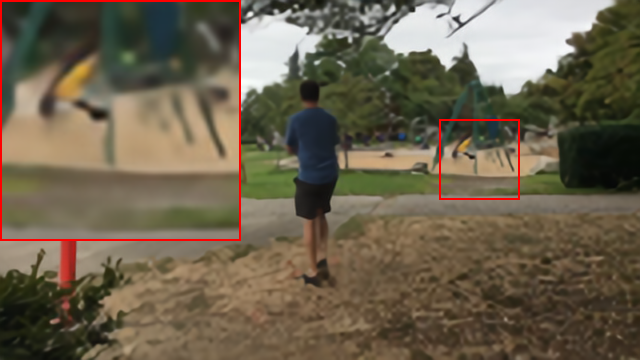}}\\
			\intervspace
			{\includegraphics[width=\linewidth]{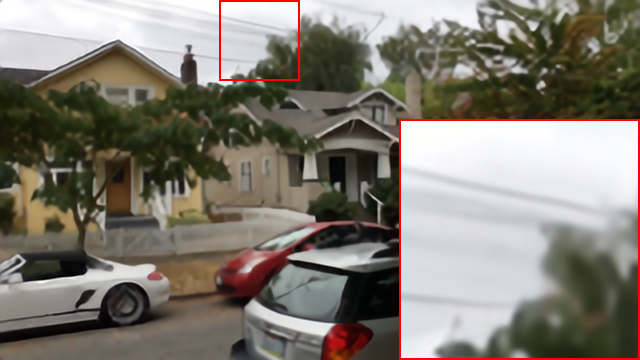}}\\
			\intervspace
			{\includegraphics[width=\linewidth]{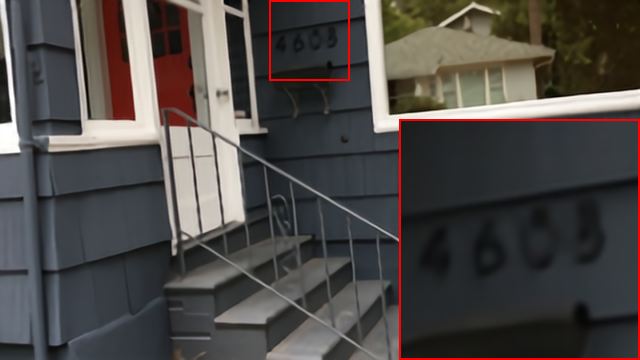}}\\
			\intervspace
			\intervspace
			\centerline{BMBC+MetaSR}
		\end{minipage}
		\begin{minipage}[htb]{0.16\linewidth}
			\centering
			{\includegraphics[width=\linewidth]{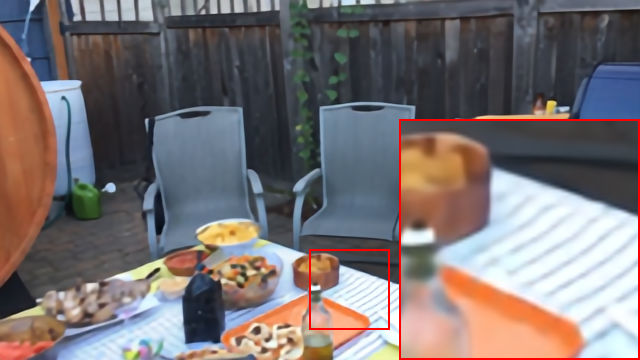}}\\
			\intervspace
			{\includegraphics[width=\linewidth]{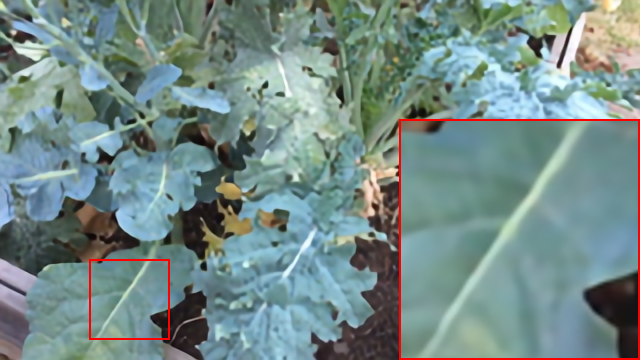}}\\
			\intervspace
			{\includegraphics[width=\linewidth]{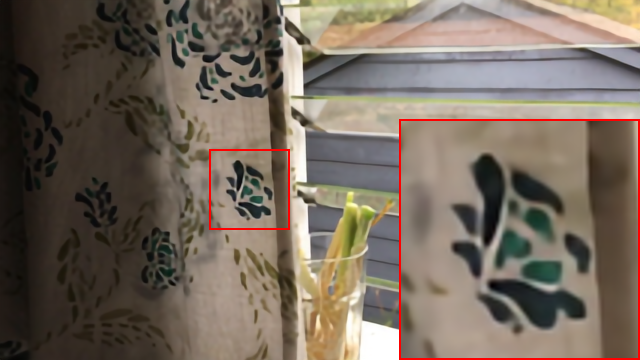}}\\
			\intervspace
			{\includegraphics[width=\linewidth]{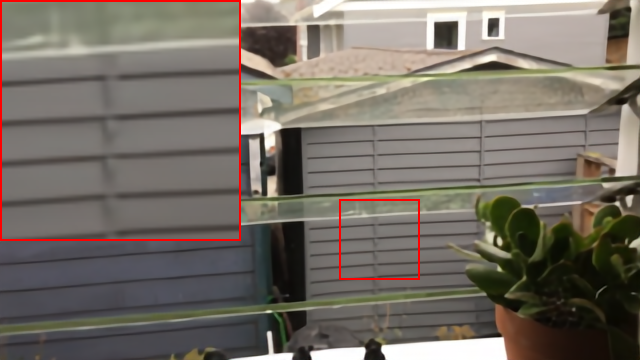}}\\
			\intervspace
			{\includegraphics[width=\linewidth]{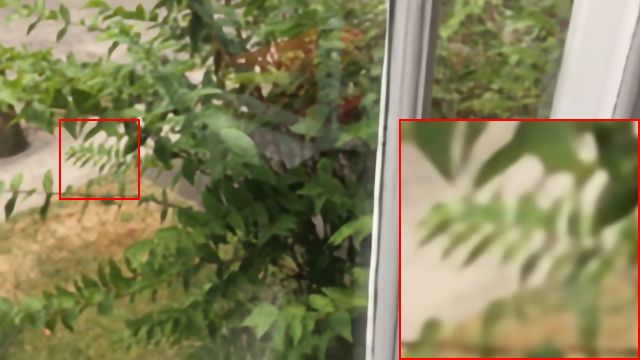}}\\
			\intervspace
			{\includegraphics[width=\linewidth]{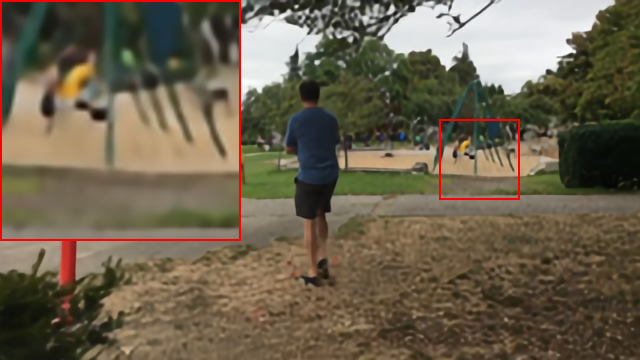}}\\
			\intervspace
			{\includegraphics[width=\linewidth]{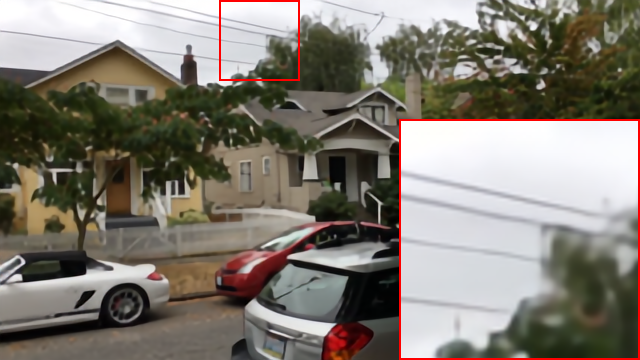}}\\
			\intervspace
			{\includegraphics[width=\linewidth]{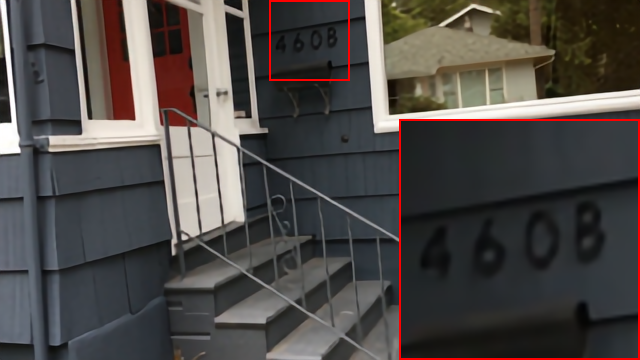}}\\
			\intervspace
			\intervspace
			\centerline{USTVSRNet}
		\end{minipage}
		\begin{minipage}[htb]{0.16\linewidth}
			\centering
			{\includegraphics[width=\linewidth]{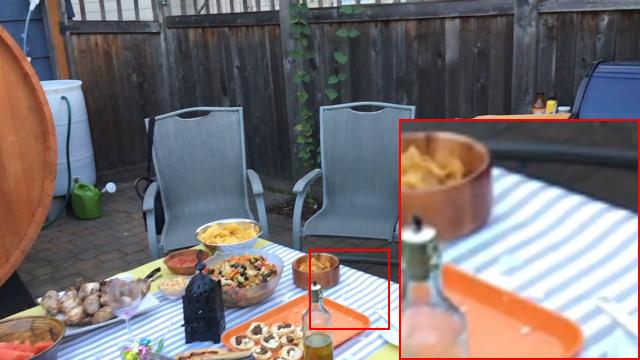}}\\
			\intervspace
			{\includegraphics[width=\linewidth]{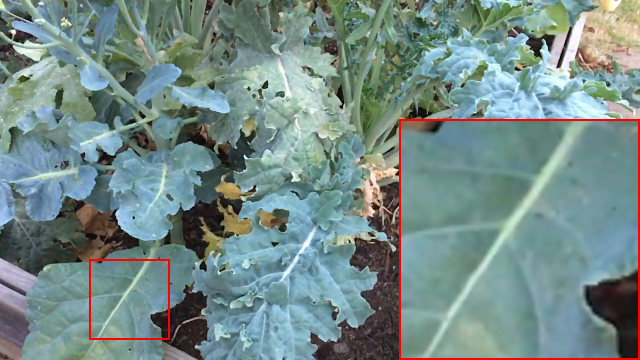}}\\
			\intervspace
			{\includegraphics[width=\linewidth]{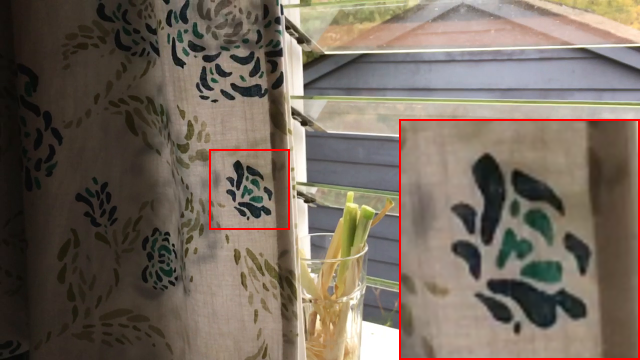}}\\
			\intervspace
			{\includegraphics[width=\linewidth]{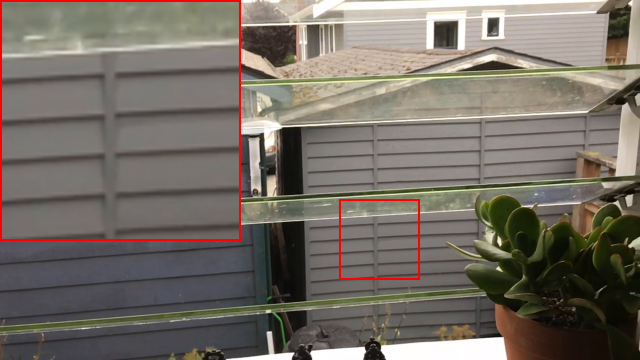}}\\
			\intervspace
			{\includegraphics[width=\linewidth]{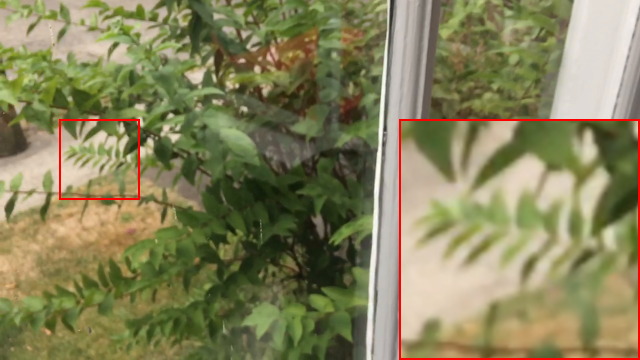}}\\
			\intervspace
			{\includegraphics[width=\linewidth]{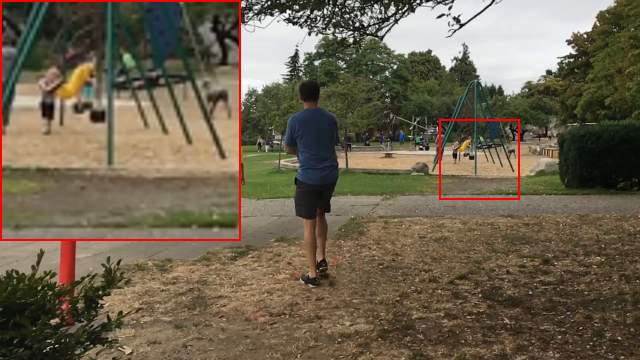}}\\
			\intervspace
			{\includegraphics[width=\linewidth]{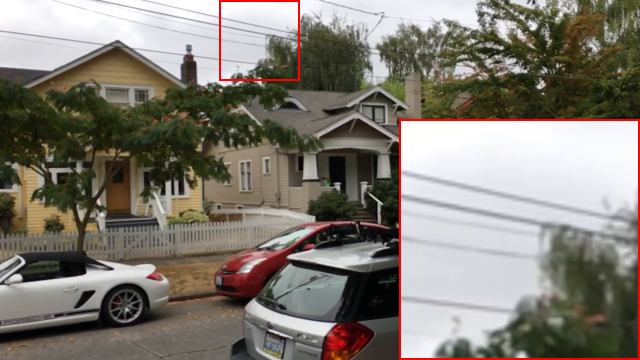}}\\
			\intervspace
			{\includegraphics[width=\linewidth]{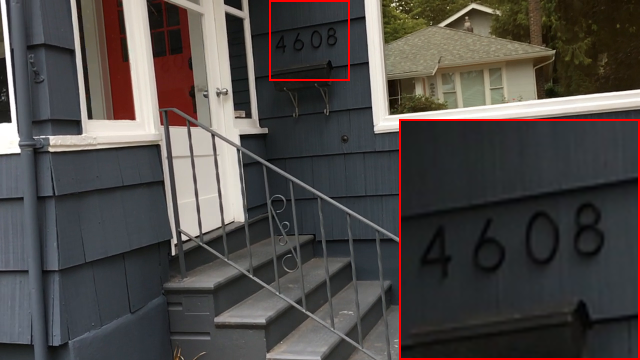}}\\
			\intervspace
			\intervspace
			\centerline{GT}
		\end{minipage}
	\end{center}
	\figvspace
	\caption{Qualitative comparisons of different Unconstrained STVSR algorithms.}
	\label{fig:qualitative comparisons uncon}
\end{figure*}

\noindent \textbf{Perceptual Loss:}
Different from the per-pixel loss, the perceptual loss seeks to measure the difference from a global visual view, which has been shown effective in generating visually realistic images. The perceptual loss often leverages multi-scale feature maps extracted from a pre-trained network to quantify the difference. Here, we adopt VGG-16 \cite{simonyan2014very} as the pre-trained network, and use feature maps from the last layer of each of the first three stages to measure the difference (\textit{i.e.}, Conv1\_2, Conv2\_2 and Conv3\_3). The loss can be expressed in the following form:
\begin{equation}
\begin{aligned}
\mathcal{L}_p = \sum_{i=1}^{3}||\Phi_i (\hat{I}_t^H)-\Phi_i (I_t^H)||_2^2,
\end{aligned}
\end{equation}
where $\Phi_i (I_t^H), i=1, 2, 3$ are the aforementioned three feature maps corresponding to $I_t^H$ while $\Phi_i (\hat{I}_t^H)$ corresponds to $\hat{I}_t^H$.

\noindent \textbf{Overall Loss:}
By combining the $L_1$ loss and the perceptual loss, the overall loss can be defined as:
\begin{equation}
\begin{aligned}
\mathcal{L} = \mathcal{L}_1 + \lambda \mathcal{L}_p,
\end{aligned}
\end{equation}
where $\lambda$ is a hyper-parameter to balance the $L_1$ loss term and the perceptual loss term. Experimentally, we find setting $\lambda=0.04$ reaches the best performance.

\subsubsection{Training Dataset}
Adobe-240 dataset \cite{su2017deep} consists of $133$ handheld recorded videos, which mainly contain outdoor scenes. The frame rate of each video is $240$ fps, with spatial resolution as $720 \times 1,280$. From this set, $103$ videos are randomly selected to construct our training dataset. That set is formed by successively grouping every $9$ consecutive frames, and resizing them to $360 \times 640$ to form a training sequence $I_0^H, I_{0.125}^H, \cdots, I_1^H$. In this way, we obtain $10,895$ sequences in total.
The LR frames are generated through bicubic down-sampling from the HR frames. We randomly crop image patches of size $56 \times 56$ from the LR frames for training. Horizontal/vertical flipping as well as temporal order reversal is performed for data augmentation.

\subsubsection{Training Strategy}
During the training phase, $t$ and $s$ are randomly selected to build each training batch. The image patches within a single batch share the same $t$ and $s$. We adopt the Adam optimizer \cite{kingma2014adam} with a batch size of $18$, where $\beta_1$ and $\beta_2$ are set to the default values $0.9$ and $0.999$, respectively. We train our network for $30$ epochs in total with the initial learning rate set to $10^{-4}$, and the learning rate is reduced by a factor of $10$ at epoch $20$. The training is carried out on two NVIDIA GTX 2080Ti GPUs, which takes about one day to converge.

\subsection{Evaluation Dataset}
\subsubsection{Adobe Testing Dataset \cite{su2017deep}}
We treat the remaining $30$ videos of the Adobe-240 dataset as an evaluation dataset. As in the case of the training dataset, we successively group every $9$ consecutive frames (resized to $360 \times 640$), resulting in $2,560$ test sequences. For each sequence, the LR frames are generated from the HR ones via bicubic down-sampling.

\subsubsection{Gopro Testing Dataset \cite{nah2017deep}}
This dataset contains $11$ videos recorded by a hand-held camera. The frame rate of each video is $240$ fps, and the image resolution is $720 \times 1,280$. The dataset is released in image format with a total of $12,221$ images. We successively group every $9$ consecutive images as a test sequence. In this way, $1,355$ test sequences are generated.

\subsection{Comparisons to SOTA methods}
To the best of our knowledge, there is no one-stage method of this kind in the literature. So we only consider two-stage methods composed of SOTA unconstrained VFI methods (BMBC \cite{park2020bmbc} and DAIN \cite{bao2019depth}) and SOTA SISR methods (since the code of \cite{wang2020learning} is not publicly available, we choose to use Meta-SR \cite{hu2019metasr}). Here we set $t={0, 0.125, \cdots, 1}$ and $s={1, 1.5, \cdots, 4}$ respectively.

\begin{table}[h]
	\resizebox{\textwidth}{!}{
		\begin{tabular}{l|cc}
			\toprule
			Method   & \hspace{5mm}\#Parameters (M)\hspace{5mm} &\hspace{5mm}Runtime (s)\hspace{5mm} \\
			\hline
			USTVSRNet	&$12$ &$0.12$\\
			\hline
			BMBC + Meta-SR &$33$&$0.34$\\
			BMBC + Bicubic &$11$&$0.21$\\
			\hline
			DAIN+Meta-SR & $46$&$0.35$\\
			DAIN+Bicubic & $24$&$0.18$\\
			\bottomrule
	\end{tabular}}
	\caption{Model size and running time comparisons with $s=4$, where the model size is reported in millions (M) and the running time is reported in second (s) per frame.}
	\label{tab:arb cost}
\end{table}

Fig.~\ref{fig:quantitative comparsion adobe} and Fig.~\ref{fig:quantitative comparsion gopro} show (a)-(c) PSNR scores for different temporal positions with $s=2.5, 3.5, 4.0$, (d)-(f) PSNR scores for different scale factors with $t=0.375, 0.500, 0.625$, on the Adobe240 and Gopro testing datasets, respectively. More results can be found in the supplementary materials.
According to the experimental results, we make two observations: 1) the usage of more advanced VFI or SR methods contribute to better predicted results. For example, although BMBC+MetaSR and DAIN+MetaSR are both equipped with the same SR method, the former performs better than the latter due to the fact that BMBC is more advanced than DAIN; 2) compared to two-stage methods, the proposed method has better performance and is much more stable. This is due to the fact that the components of the two-stage methods work in isolation and cannot exploit the relationships between behavior in space and time.

Fig.~\ref{fig:qualitative comparisons uncon} shows some qualitative comparisons with $t=0.5$ and $s=4$. It can be seen that the proposed method tends to generate more visually appealing results than the others.
For instance, the proposed USTVSRNet yields sharper and clearer strips in the first row of Fig.~\ref{fig:qualitative comparisons uncon}; the leaves and the flower pattern generated by our method are much clearer than others in the second and the third rows, respectively.

Tab.~\ref{tab:arb cost} provides comparisons between these methods in terms of model size and running time. Since two-stage methods are simple concatenations of the VFI and SR algorithms, they tend to be overweight and slow. In contrast, the proposed method is more compact and efficient. Specifically, compared to the best performing two-stage method, namely BMBC + Meta-SR, the proposed USTVSRNet only has about $1/4$ of size and takes half time to reconstruct one frame.

\subsection{Ablation Study}


\begin{table}[h]
	\caption{Quantitative results of ablation study regarding FINet and EnhanceNet with $s=1, 2, 3, 4$, where PSNR and SSIM scores are averaged over $t$.}
	\label{tab:ab1}
	\footnotesize
	\center
	\renewcommand\tabcolsep{3.0pt}
	\resizebox{\textwidth}{!}{\begin{tabular}{ccccccccc}
			\toprule
			\multicolumn{1}{c}{\multirow{2}*{Method}} 
			&\multicolumn{2}{c}{$s=1$}
			&\multicolumn{2}{c}{$s=2$}
			&\multicolumn{2}{c}{$s=3$}
			&\multicolumn{2}{c}{$s=4$}\\ 
			\cmidrule(r){2-3} 
			\cmidrule(r){4-5}
			\cmidrule(r){6-7}
			\cmidrule(r){8-9}
			& PSNR & SSIM & PSNR & SSIM & PSNR & SSIM & PSNR & SSIM\\
			\hline
			(a) & $31.44$& $0.9140$ & $30.09$& $0.8949$& $27.82$& $0.8357$& $26.42$& $0.7767$\\
			(b) & $32.02$& $0.9249$ & $29.81$& $0.8882$& $27.38$& $0.8182$& $25.99$& $0.7550$\\
			(c) & $32.39$& $0.9270$ & $30.85$& $0.9078$& $28.44$& $0.8542$& $26.83$& $0.7945$\\
			\bottomrule
			
	\end{tabular}}
	
\end{table}

\begin{figure}[h]
	\centering
	\begin{minipage}[b]{0.45\linewidth}
		\centering
		\includegraphics[width=\linewidth]{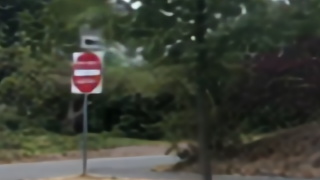}
		\scriptsize{(a) $t=0.25$ w/o FINet}
	\end{minipage}
	\begin{minipage}[b]{0.45\linewidth}
		\centering
		\includegraphics[width=\linewidth]{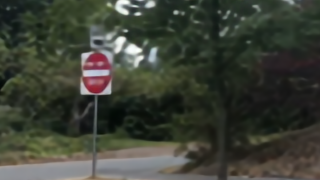}
		\scriptsize{(b) $t=0.75$ w/o FINet}
	\end{minipage}
	\\
	\begin{minipage}[b]{0.45\linewidth}
		\centering
		\includegraphics[width=\linewidth]{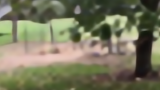}
		\scriptsize{(c) w/o EnhanceNet}
	\end{minipage}
	\begin{minipage}[b]{0.45\linewidth}
		\centering
		\includegraphics[width=\linewidth]{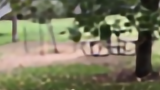}
		\scriptsize{(d) w/ EnhanceNet}
	\end{minipage}
	\figvspace
	\caption{Qualitative results of ablation study regarding FINet and EnhanceNet.}
	\label{fig:ab1}
\end{figure}

\subsubsection{Effectiveness of image-level and feature-level interpolation}

To validate the effectiveness of the image-level and feature-level interpolation, we consider the following three variants: a) USTVSRNet without FINet (where the flow estimation and reverse network are still preserved to generate optical flows for EnhanceNet). For this variant, we directly pass $F_t^{'L}$ to the reconstruction network; b) USTVSRNet without EnhanceNet. For this variant, $F_t^{L}$ is directly fed into the reconstruction network; c) the complete network. We only provide the average scores over $t$ with $s=1,2, 3, 4$ in Tab.~\ref{tab:ab1} due to paper space limitation (Noted we excluded $t=0, 1$ for $s=1$). However, we observe similar results for different values of $s$.

From Tab.~\ref{tab:ab1} (a) and (c), we can make the following two observations: 1) interpolating at the image level in addition to the feature level does contribute positively to the final reconstruction. It is expected that the performance can be improved further if more advanced VFI methods are adopted; 2) even without the explicit image-level prediction by FINet, the network still retains some, albeit reduced, ability to generate intermediate frames for different target times, owing to the implicit feature-level synthesis in EnhanceNet. A visual example is illustrated in Figs.~\ref{fig:ab1} (a-b), in which two frames for different times are generated through USTVSRNet without FINet.

As can be seen from Tab.~\ref{tab:ab1} (b-c), removing EnhanceNet degrades the performance in terms of PSNR and SSIM. Indeed, in addition to the loss of feature-level interpolation, removing EnhanceNet deprives the system of the ability to refine the reference frame and gain information from the source frames for reconstruction, which leads to unsatisfactory results, as shown in Figs.~\ref{fig:ab1} (c-d).

\subsubsection{Effectiveness of GPL and SARDB}

We next demonstrate the effectiveness of the GPL and SARDB. Since the reconstruction network is functionally orthogonal to the other three sub-networks, we repurpose it as an SISR network. The following experiments are based on the SISR network and the RDN \cite{zhang2018residual} is adopted as the backbone. The Vimeo90K dataset is commonly used in the SR area, thus we adopt this dataset in this part. We first demonstrate the effectiveness of the GPL and then SARDB.

\noindent \textbf{Comparison with SPL:}
As we stated in the previous section, the proposed GPL is a generalized version of the SPL. Here we compare them in terms of the fixed scale up-sampling. The baseline is RDN, which employs the SPL at the end of the network to upscale features by a fixed scale factor. We will denote this system by S-RDN. Then, we replace the SPL by the proposed GPL to obtain a system called G-RDN.
We evaluate each method on up-sampling factors $r=2, 3, 4$, respectively. For each scale factor, the baseline RDN needs to be modified and re-trained. In contrast, for G-RDN, there is no need  to modify the network structure. For fair comparisons, we also train it on each scale factor. The experiment results can be found in Tab.~\ref{tab con}.

\begin{table}[h]
		\begin{tabular}{l|cccc}
			\toprule
			Method & scale-factor & PSNR & SSIM &Runtime\\
			\hline
			S-RDN & $\times 2$ & $40.81$ & $0.9780$ & $0.13$\\
			G-RDN & $\times 2$ & $\mathbf{40.84}$ & $\mathbf{0.9780}$ & $0.15$\\
			\hline
			S-RDN & $\times 3$ & $36.27$ & $0.9465$ & $0.07$\\
			G-RDN & $\times 3$ & $\mathbf{36.30}$ & $\mathbf{0.9468}$ & $0.09$\\
			\hline
			S-RDN & $\times 4$ & $33.88$ & $\mathbf{0.9162}$ & $0.05$\\
			G-RDN & $\times 4$ & $\mathbf{33.90}$ & $0.9161$ & $0.07$\\
			\bottomrule
		\end{tabular}
	\caption{Comparisons between SPL and GPL in terms of the fixed scale up-sampling on the Vimeo90k dataset.}
	\label{tab con}
\end{table}

As we can see from Tab.~\ref{tab con}, G-RDN achieves slightly better results than S-RDN at all scales with a negligible running time increase (except for $\times 4$ up-sampling on which G-RDN performs a little worse in terms of the SSIM value), which implies the performance of the GPL is on par or marginally better than that of the SPL in the scenario of fixed scale up-sampling. More importantly, compared to the SPL, the GPL enables the network to have the capability to perform unconstrained up-sampling task (will be demonstrated in the next), instead of restricted to certain specific scaling factors. Therefore, from these two aspects, the GPL can be considered as a generalized version of the SPL.

\noindent \textbf{Comparison with unconstrained up-sampling methods:}
Different from constrained counterparts, unconstrained methods are able to upscale an image by an arbitrary factor within a single model.
For this part, we compare the GPL with certain unconstrained up-sampling modules. Since only a few methods concentrate on arbitrary scale factor upsampling, we need to design several baseline systems. The following three baselines are taken into consideration: 1) the first baseline directly adopt the bicubic interpolation technique to up-sample images, denoted as Bicubic; 2) we first use a standard RDN to up-sample a image by $k$ times ($k$ is a fixed integer), then resize the up-sampled image to the desired size with bicubic interpolation, denoted as I-RDN($\times k$); 3) we replace the SPL of RDN with the bicubic interpolation method, which means that bicubic interpolation is used to upscale the feature maps, denoted as Bi-RDN. In addition, we also compare with the SOTA unconstrained up-sampling method, Meta-RDN \cite{hu2019metasr}, which is the same as Bi-RDN and G-RDN except for the final up-sampling module.
Bi-RDN, Meta-RDN, and G-RDN are trained with the same unconstrained training strategy.
Tab.~\ref{tab:ab uncon} shows the evaluation results, which are averaged over $s\in[1,4]$. 

\begin{table}[h]
	\begin{tabular}{c|ccc}
		\toprule
		Method & \hspace{3mm}PSNR\hspace{3mm} & \hspace{3mm}SSIM\hspace{3mm} &\hspace{3mm}Runtime\hspace{3mm} \\
		\hline
		Bicubic &$37.41$ &$0.9282$ & $0.02$\\
 		I-RDN($\times 2$) & $38.25$& $0.9548$& $0.16$\\
 		I-RDN($\times 4$) & $38.02$& $0.9578$ &$0.17$\\
		Bi-RDN & $40.37$& $0.9586$& $0.17$\\
		Meta-RDN & $40.72$& $0.9578$ & $0.20$\\
 		G-RDN & $\mathbf{40.90}$& $\mathbf{0.9590}$ & $0.18$\\
		\bottomrule
	\end{tabular}
	\caption{Comparisons for different unconstrained upscale methods on Vimeo90k dataset.}
	\label{tab:ab uncon}
\end{table}

Tab.~\ref{tab:ab uncon} illustrates the effectiveness of the proposed GPL. It has a clear advantage over other methods in terms of PSNR and SSIM. 
In particular, GPL outperforms the SOTA up-sampling module, Meta-Upscale, by about $0.18$ dB and has a faster running speed.

\begin{figure}[h]
	\centering
	\begin{minipage}[b]{0.45\linewidth}
		\centering
		{\includegraphics[width=\linewidth]{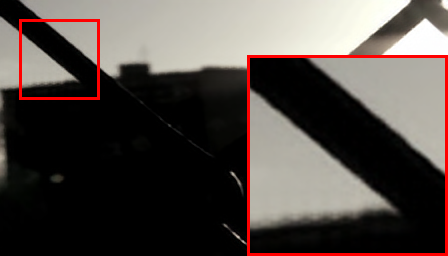}}\\			\intervspace
		{\includegraphics[width=\linewidth]{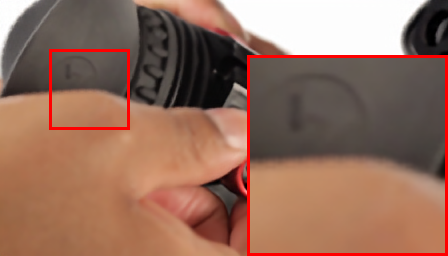}}
		\scriptsize{(a) FG-RDN}
	\end{minipage}
	\begin{minipage}[b]{0.45\linewidth}
		\centering
		{\includegraphics[width=\linewidth]{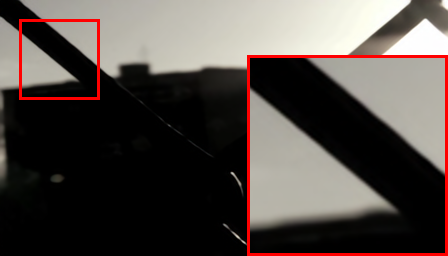}}\\			\intervspace 
		{\includegraphics[width=\linewidth]{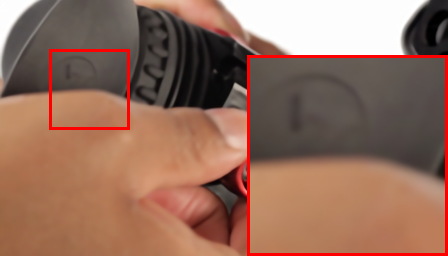}}
		\scriptsize{(b) G-RDN}
	\end{minipage}
	\figvspace
	\caption{Qualitative results of ablation study regarding channel dimension freedom.}
	\label{fig:channel freedom}
\end{figure}

\begin{table*}[t]
	\caption{Quantitative comparisons on Adobe240 dataset, Gopro dataset, and Vimeo90K dataset with $s=4$, where `Center' denotes the PSNR and SSIM value of the center frame $\hat{I}_{0.5}^H$ while `Whole' is for average values of $\hat{I}_{0}^H$, $\hat{I}_{0.5}^H$, and $\hat{I}_{1}^H$. We highlight the first place and the second place in red and blue, respectively.}
	\footnotesize
	\center
	\renewcommand\tabcolsep{7.0pt}
	\resizebox{\textwidth}{!}{\begin{tabular}{ccccccccc}
			\toprule
			\multicolumn{1}{c}{\multirow{2}*{Method}} 
			&{\#Parameters} 
			&\multicolumn{2}{c}{Adobe240} 
			&\multicolumn{2}{c}{Gopro}
			&\multicolumn{2}{c}{Vimeo90k} \\ 
			\cmidrule(r){3-4} 
			\cmidrule(r){5-6}
			\cmidrule(r){7-8}
			& (million) & Center & Whole & Center & Whole & Center & Whole\\
			\hline
			USTVSRNet&$12$ &{\color{red}$26.64/0.7944$} & {\color{red}$27.43/0.8102$} & {\color{red}$25.94/0.8093$} & {\color{red}$26.91$}{\color{blue}$/0.8347$} & {\color{red}$30.32/0.8866$} & {\color{red}$31.20/0.8956$}\\
			TMNet &$12$& {\color{blue}$26.38/0.7890$}&{\color{blue}$27.28/0.8086$} & {\color{blue}$25.88/0.8083$}& {\color{blue}$26.90$}{\color{red}$/0.8352$}& {\color{blue}$30.20/0.8847$}& {\color{blue}$31.09/0.8942$}&\\
			Zooming SloMo &11 & $26.36/0.7874$& $27.26/0.8069$& $25.84/0.8066$& $26.85/0.8333$& $30.11/0.8830$& $31.02/0.8928$\\
			STARNet &$58$ & $26.01/0.7639$ & $27.01/0.7895$ & $25.80/0.8053$ & $26.80/0.8324$ & $30.11/0.8828$ & $30.92/0.8907$ \\
			\hline
			AdaCoF+RSDN &$55$ &$25.77/0.7583$ &$26.90/0.7888$ &$25.27/0.7825$ &$26.46/0.8187$ & $29.34/0.8654$ & $30.69/0.8843$\\
			AdaCoF+RBPN &$35$ &$25.68/0.7501$ &$26.64/0.7766$ &$25.27/0.7773$ &$26.25/0.8074$ & $29.32/0.8612$ & $30.48/0.8781$\\
			AdaCoF+DBPN &$32$ &$25.44/0.7322$ &$26.46/0.7630$&$25.03/0.7612$& $26.01/0.7945$&$29.13/0.8543$&$30.30/0.8732$\\
			\hline
			BMBC+RSDN &$44$ &$25.82/0.7575$ &$26.95/0.7883$ &$25.30/0.7808$ &$26.54/0.8195$ & $29.46/0.8670$& $30.75/0.8849$\\
			BMBC+RBPN &$24$ &$25.73/0.7504$ &$26.67/0.7767$ &$25.29/0.7762$ & $26.26/0.8069$ & $29.36/0.8636$ & $30.50/0.8793$\\
			BMBC+DBPN &$32$ &$25.51/0.7329$ &$26.50/0.7633$ &$25.07/0.7608$ & $26.03/0.7643$ & $29.28/0.8568$ & $30.38/0.8745$\\
			\bottomrule
	\end{tabular}}
	\label{tab:comparison with SOTA}
\end{table*}

\begin{table}[h]
	\begin{tabular}{l|cccc}
		\toprule
		Method & dimension & PSNR & SSIM &Runtime\\
		\hline
		FG-RDN & $5\times64$ &$39.37$ &$0.9525$ &$0.17$\\
		\hline
		G-RDN & $1\times64$ & $40.78$ & $0.9587$ & $0.18$\\
		G-RDN & $3\times64$ & $40.86$ & $0.9589$ & $0.18$\\
		G-RDN & $5\times64$ & $\mathbf{40.90}$ & $\mathbf{0.9590}$ & $0.18$\\
		G-RDN & $7\times64$ & $40.90$ & $0.9590$ & $0.18$\\
		\bottomrule
	\end{tabular}
	\caption{ Comparisons of FG-RDN and G-RDN with different $C_\text{mid}$.
		\label{tab:dimension}}
\end{table}

\noindent \textbf{Importance of channel direction freedom:} To illustrate the importance of channel direction freedom, we consider the following variant: we keep other components the same as G-RDN except for forcing $\triangle p_c=0$, in which the sampling position cannot move along the channel direction. This system is denoted by FG-RDN.
The experimental results can be found in Tab.~\ref{tab:dimension}, where one can easily find freedom of channel direction that leads to better reconstruction results. Indeed, if $\Delta p_c=0$, then the local feature vectors of the output feature maps will become identical, which tends to generate blurry results or jagged edges. A visual example can be found in Fig.~\ref{fig:channel freedom}. Specifically, consider two different output positions on the output feature maps ($i_1, j_1$) and ($i_2,j_2$). If they are projected to the same location on the intermediate feature maps $\lfloor{i}'_1\rfloor, \lfloor{j}_1'\rfloor=\lfloor{i}_2'\rfloor, \lfloor{j}_2'\rfloor$, their output feature vectors $\operatorname{GPL}(T)_{i_1,j_1}$ and $\operatorname{GPL}(T)_{i_2,j_2}$ will be exactly the same according to Eq.~(\ref{eq:gpl}) (due to $\Delta p_c=0$), which limits the diversity of the output feature maps. From another point of view, forcing $\triangle p_c=0$ corresponds to using the nearest interpolation to up-sample the feature maps. Naturally, its performance is not as good as that of G-RDN, and it is even worse than Bi-RDN (since bicubic interpolation is superior to nearest interpolation in nature). Therefore, the channel direction freedom plays an important role in the GPL.

\noindent \textbf{Choice of $C_\text{mid}$:}
For the all experiments above, we set $C_\text{mid} = 5C_\text{in} = 5C_\text{out}=5\times64$. Now, we investigate how to choose the channel dimension of intermediate feature maps $C_\text{mid}$. We fix $C_\text{in}=C_\text{out}=64$ and vary $C_\text{mid}$. Tab.~\ref{tab:dimension} shows the results, which are averaged over $s\in[1,4]$.
As shown in Tab.~\ref{tab:dimension}, as the dimension increases, the performance improves initially, but eventually becomes saturated. In particular, setting $C_\text{mid}$ to more than $7 \times 64$ does not further improve the quality of the reconstructed HR image.

\noindent \textbf{Effectiveness of SARDB:}
We finally investigate the contribution of the SARDB. Two networks are trained and evaluated on scale factor $r \in [1, 4]$: one with SARDB; the other one with RDB.
We experimentally find that the scale-dependent features generated by SARDB improve the performance by $0.28$ dB and $0.0012$ in terms of PSNR and SSIM, respectively, with negligible increasing in the running cost.

\section{Experiments For Fixed Space-Time Video Super-Resolution}
\label{sec:fix}
Different from unconstrained STVSR, in fixed STVSR the temporal frame rate and the spatial resolution are not adjustable without retraining or modifying the network. Some experimental comparisons with  fixed STVSR are provided below.

\subsection{Implementation Details}
In this section, $t$ can only vary among $\{0, 0.5, 1\}$ and $s$ is set to $4$, which means the network can only up-sample a video by $\times2$ and $\times 4$ times in terms of temporal and spatial resolutions, respectively. As in the previous section, we set $K=4$ and $C_\text{mid} = 5C_\text{in} = 5C_\text{out}=5\times64$. The training dataset and training strategy are described below.

\subsubsection{Training Dataset}
Same as \cite{haris2020space}, the Vimeo90k Triplet Training Dataset \cite{xue2019video} is adopted to train our model, where we have $51,312$ sequences in total and the image resolution is $256 \times 448$. Within each sequence, the first, the second and the third frames are treated as $I_0^H$, $I_{0.5}^H$ and $I_1^H$, respectively. Similarly, we use the bicubic down-sampling method to generate LR images from HR ones. We also perform horizontal and vertical flips, as well as temporal order reversal, for data augmentation.

\subsubsection{Training Strategy}
For each training iteration, $t$ is randomly selected from $\{0, 0.5, 1\}$ and $s$ is set as $4$ to construct the corresponding training batch. The Adam optimizer is adopted with a batch size of $24$. We train the network for $25$ epochs in total, with the initial learning rate as $10^{-4}$. The learning rate is reduced by $\times 2$ times at every $8$ epochs for the first $16$ epochs and by $\times 5$ times every $3$ epochs for the last $9$ epochs.
The training is carried out on two NVIDIA GTX 2080Ti GPUs, which takes about one day to converge.

\subsection{Evaluation Dataset}
\subsubsection{Adobe Testing Dataset and Gopro Testing Dataset}
The Adobe Testing Dataset and the Gopro Dataset from the previous section are directly used as the two evaluation datasets for this section. There are $2,560$ and $1,355$ sequences in the Adobe dataset and the Gopro dataset, respectively, each with $9$ frames. We only use the first, the fifth and the last frame of each sequence to compare different algorithms.

\subsubsection{Vimeo90k Triplet Testing Dataset \cite{xue2019video}}
This dataset consists of $3,782$ video sequences, each with $3$ frames. The image resolution of this dataset is $256 \times 448$. The first, second, and third frames in each video sequence are treated as $I_0^H$, $I_{0.5}^H$, and $I_1^H$ respectively.

\subsection{Comparisons to SOTA Methods}
Here the STARNet \cite{haris2020space}, Zooming SloMo \cite{xiang2020zooming}, and TMNet \cite{xu2021temporal} are chosen as representatives of one-stage fixed STVSR methods. For fair comparison, they are retrained on our training dataset using the same strategy. As to two-stage methods, we combine pre-trained SOTA VFI methods (AdaCoF \cite{lee2020adacof} and BMBC \cite{park2020bmbc}) and SR methods (RSDN \cite{isobe2020video}, RBPN \cite{haris2019recurrent} and DBPN \cite{haris2018deep} are chosen as representatives of the VSR and SISR methods, respectively).

We quantitatively compare our method with the chosen one-stage and two-stage methods under two well-known objective image quality metrics (PSNR and SSIM). The scores of the center frame and the average scores over all three frames are provided in Tab.~\ref{tab:comparison with SOTA}. 
It can be seen that the proposed method ranks consistently at the top performance-wise, and comes in a close second in terms of the number of parameters. The two-stage methods not only suffer from large model size, but also lack competitiveness in performance since the constituent VFI and SR techniques are constrained to work in isolation. Although STARNet, Zooming SloMo, and TMNet are capable of handling diverse space-temporal patterns and improve the performance significantly, they are still behind the proposed method by a visible gap.

\section{Conclusion}
In summary, we have proposed an unconstrained STVSR method that has the freedom
to arbitrarily adjust the temporal frame rate and spatial resolution of the output video. Beyond using the optical flow technique for temporal interpolation, several new ideas are introduced, which include the generalized pixelshuffle operation for upsampling, 
a refined mechanism to generate scale-adaptive features, and the integration of image-level and feature-level representations. Despite their excellent performance, it is conceivable that these new ideas could be further developed to yield even better performance. Moreover, there could well be likely alternative approaches to realizing unconstrained STVSR. In this sense, our work should be viewed as a stepping-stone towards a full-fledged framework for AI-enabled STVSR.

\ifCLASSOPTIONcaptionsoff
\newpage
\fi

\bibliographystyle{IEEEtran}
\bibliography{IEEEabrv,egbib}
\end{document}